\documentclass[sigconf]{aamas} 

\usepackage{afterpage}
\usepackage{tcolorbox}

\usepackage{balance}

\title[Pavlovian Signalling with General Value Functions]{Pavlovian Signalling with General Value Functions in Agent-Agent Temporal Decision Making}

\author{Andrew Butcher}
\affiliation{%
  \institution{DeepMind Technologies, Ltd.}
  \streetaddress{10065 Jasper Avenue, Suite 250}
  \city{Edmonton}
  \state{Alberta}
  \country{Canada}
  \postcode{T5J 3B1}
}
\email{abutcher@deepmind.com}

\author{Michael Bradley Johanson}
\affiliation{%
  \institution{DeepMind Technologies, Ltd.}
  \streetaddress{10065 Jasper Avenue, Suite 250}
  \city{Edmonton}
  \state{Alberta}
  \country{Canada}
  \postcode{T5J 3B1}
}

\author{Elnaz Davoodi}
\affiliation{%
  \institution{DeepMind Technologies, Ltd.}
  \streetaddress{10065 Jasper Avenue, Suite 250}
  \city{Edmonton}
  \state{Alberta}
  \country{Canada}
  \postcode{T5J 3B1}
}

\author{Dylan J. A. Brenneis}
\orcid{0000-0002-6359-0832}
\affiliation{%
  \institution{DeepMind Technologies, Ltd.}
  \streetaddress{10065 Jasper Avenue, Suite 250}
  \city{Edmonton}
  \state{Alberta}
  \country{Canada}
  \postcode{T5J 3B1}
}

\author{Leslie Acker}
\affiliation{%
  \institution{DeepMind Technologies, Ltd.}
  \streetaddress{10065 Jasper Avenue, Suite 250}
  \city{Edmonton}
  \state{Alberta}
  \country{Canada}
  \postcode{T5J 3B1}
}

\author{Adam S. R. Parker}
\affiliation{%
  \institution{University of Alberta}
  \city{Edmonton}
  \state{Alberta}
  \country{Canada}
}

\author{Adam White}
\affiliation{%
  \institution{DeepMind Technologies, Ltd. \& University of Alberta \& Alberta Machine Intelligence Institute}
  \city{Edmonton}
  \state{Alberta}
  \country{Canada}
  \postcode{T5J 3B1}
}

\author{Joseph Modayil}
\affiliation{%
  \institution{DeepMind Technologies, Ltd.}
  \streetaddress{10065 Jasper Avenue, Suite 250}
  \city{Edmonton}
  \state{Alberta}
  \country{Canada}
  \postcode{T5J 3B1}
}

\author{Patrick M. Pilarski}
\affiliation{%
  \institution{DeepMind Technologies, Ltd. \& University of Alberta \& Alberta Machine Intelligence Institute}
  \city{Edmonton}
  \state{Alberta}
  \country{Canada}
  \postcode{T5J 3B1}
}

\begin{abstract}
 In this paper, we contribute a multi-faceted study into {\em Pavlovian signalling}---a process by which learned, temporally extended predictions made by one agent inform decision-making by another agent. Signalling is intimately connected to time and timing. In service of generating and receiving signals, humans and other animals are known to represent time, determine time since past events, predict the time until a future stimulus, and both recognize and generate patterns that unfold in time. We investigate how different temporal processes impact coordination and signalling between learning agents by introducing a partially observable decision-making domain we call the Frost Hollow. In this domain, a prediction learning agent and a reinforcement learning agent are coupled into a two-part decision-making system that works to acquire sparse reward while avoiding time-conditional hazards. We evaluate two domain variations: machine agents interacting in a seven-state linear walk, and human-machine interaction in a virtual-reality environment. Our results showcase the speed of learning for Pavlovian signalling, the impact that different temporal representations do (and do not) have on agent-agent coordination, and how temporal aliasing impacts agent-agent and human-agent interactions differently. As a main contribution, we establish Pavlovian signalling as a natural bridge between fixed signalling paradigms and fully adaptive communication learning between two agents. We further show how to computationally build this adaptive signalling process out of a fixed signalling process, characterized by fast continual prediction learning and minimal constraints on the nature of the agent receiving signals. Our results therefore suggest an actionable, constructivist path towards communication learning between reinforcement learning agents.
\end{abstract}

         
\newcommand{\BibTeX}{\rm B\kern-.05em{\sc i\kern-.025em b}\kern-.08em\TeX}

\begin{document}

\pagestyle{fancy}
\fancyhead{}

\setcopyright{none}
\settopmatter{printacmref=false}
\renewcommand\footnotetextcopyrightpermission[1]{} 
\pagestyle{plain}
\maketitle 

\section{Introduction: Signals and reflexes}
\label{sec:intro}

Communication learning by machines promises substantial benefits when compared to hand-engineered communication systems for machine-machine or human-machine interaction \cite{lazaridou2020,crandall2018}. Despite the promise of increased flexibility, decreased human design effort, and the potential for ongoing adaptation, emergent communication learning by machines remains challenging \cite{lazaridou2020}. In this work, we explore a stepping stone between hand-engineered communication and fully learned communication. Our approach captures some of the flexibility and adaptability of machine-learned communication while also providing rapid learning and minimal assumptions about the interacting agents. Early work by Pavlov showed how signals, and signals of signals, elicit reflexive action in animals \cite{pavlov1927,windholz1990,scottphillips2014}. We propose that these reflexes also can play the role of useful signals (feature outputs) intended to inform decision-making by another agent or a component of a single agent.

 Signals carry information. In the simplest form, we can think of signals as a means of transmitting information, however the informational content and quantity of information in a signal has made signalling an area of study for philosophers, information theorists, linguists, cognitive scientists and computer scientists~\citep{dretske1981,scottphillips2009,scottphillips2014}. 
 Signalling systems are not limited to humans, but are used by all levels of biological organization. For instance, monkeys~\citep{cheney2018monkeys}, birds~\citep{charrier2005call}, bees~\citep{riley2005flight, seeley2006group} and even bacteria~\citep{taga2003chemical} have signaling systems.

An important aspect of the informational content of a signal is the symbol grounding problem---the association between symbols and their meanings \citep{vogt2007language}. Specifically in the case of classical Pavlovian conditioning, signals become associated with and play an important role in forecasting a significant event \cite{pavlov1927}: {\em a stimulus elicits a prediction that itself takes on the role of a grounded cue to help inform behaviour}. 
This relationship between learned predictions and control has been extensively studied. Unlike the instrumental learning experiments of Skinner~\citep{rescorla1988pavlovian} which required the animal to make a decision that was associated with differences in later reward, the learning in classical conditioning involves neither rewards nor decisions.  In the absence of an explicit reward maximizing mechanism for behavior change, it is natural to inquire how learning a prediction by classical conditioning can change behavior.

One answer is to connect the learned prediction of a stimulus to an unconditioned, fixed response. This approach, termed Pavlovian control, has a simple computational realization~\citep{modayil2014prediction, dalrymple2020pavlovian}.  Namely, a fixed policy emits the action $a_1$ when a stimulus $s_1$ is predicted above some threshold $\tau$, and the action $a_2$ otherwise. The ideal setting of $\tau$ depends on the timescale of the prediction, the lead time needed for the response action to be effective, and the accuracy of the prediction. Thus, Pavlovian control emits a fixed reflex-like response to the prediction of an event. This simple form of coupling the prediction to behaviour is only one of many possibilities. In the present work we study the way that predictions can be coupled to the generation of grounded signals used by other agents or components of the same agent, which we term {Pavlovian signalling}; primarily, {\em we contribute evidence as to the degree to which this signalling approach can be rapidly learned and be useful during online learning}, across domain variants and representations.


\section{Pavlovian Signalling}

Pavlovian signalling for agent-agent interaction is a process where learned, temporally extended predictions are mapped in a fixed way to signals intended for receipt by a decision-making agent, and where these signals are grounded for the sender in the definition of the predictive question. In the context of this study, a signal conveys information about the occurrence of the next event. Following the Gricean maxims of communication~\citep{grice1975logic, grice1989studies}, we focus here on a single binary {\em token} (i.e., a two-state symbol) to represent and convey a piece of information about future events.

\begin{figure}[!t]
\centering
\includegraphics[width=2.9in]{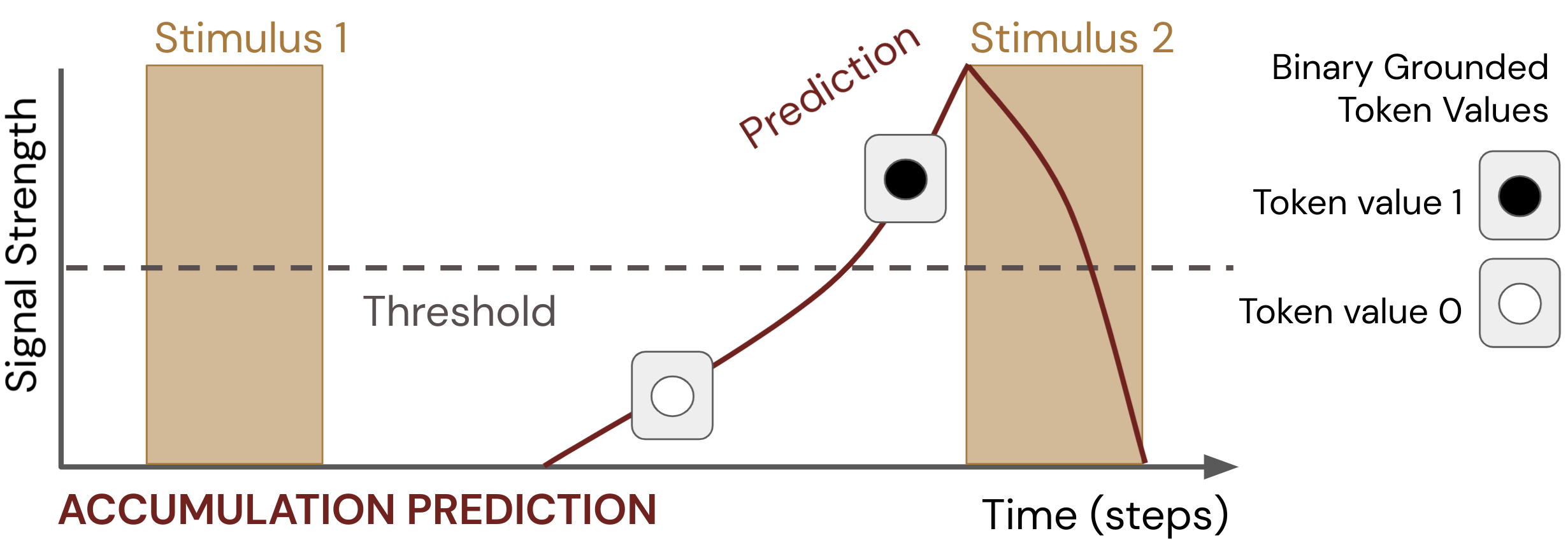}\\
(a)\\
\includegraphics[width=2.9in]{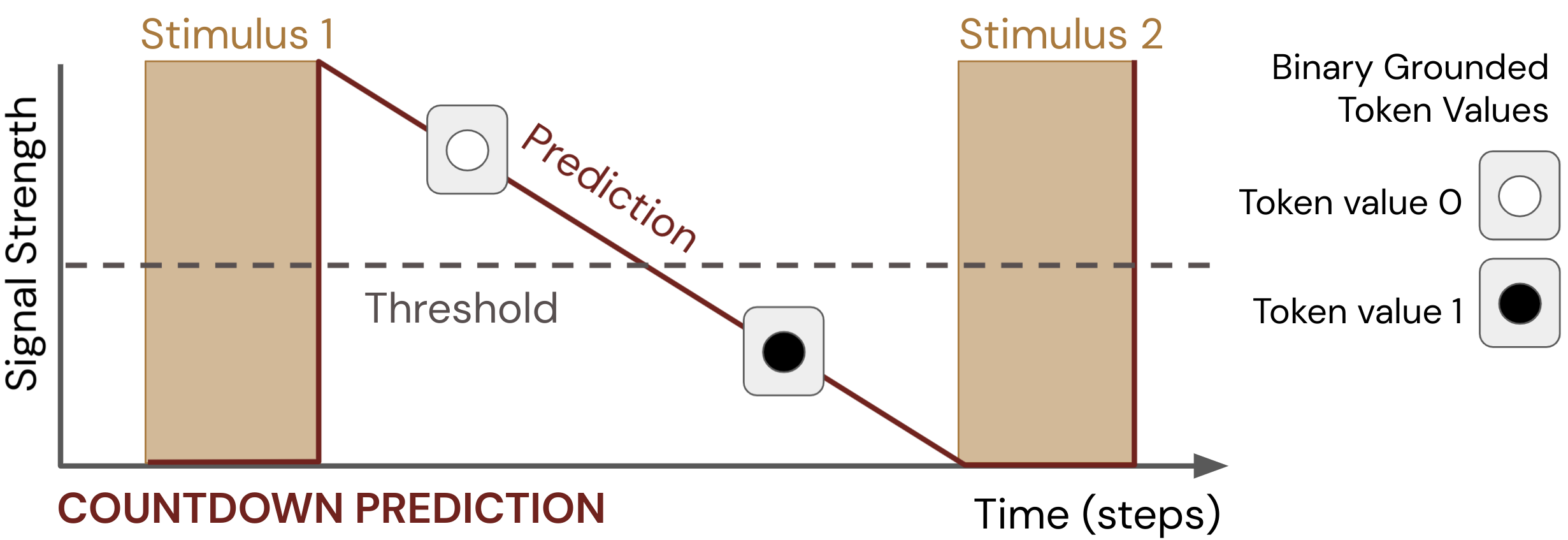}\\
(b)
\caption{{\bf Pavlovian signalling in schematic form}. Shown here are two different idealized predictions that either (a) rise in advance of an impending stimulus or (b) decrease to forecast the time until a future stimulus.}
\Description{Mechanisms for Pavlovian signalling are shown for accumulation GVF and countdown GVF questions. Two stimuli, separated in time on the x-axis are plotted with magnitude 1 on the y-axis, 

Answers to the GVF questions are mapped to binary tokens by a thresholding mechanism. The thresholding mechanism is different for both questions; the value 0 indicates no hazard is immediately threatened, and 1 indicates that a hazard is imminent.

Answers to both GVF questions are plotted over time. The answer to the Accumulation GVF question reaches a maximum at the onset of the second stimulus, and falls back to zero by the end of the second stimulus. The answer to the Countdown GVF question is at a maximum immediately following the first stimulus, and drops to 0 at the beginning of the second stimulus. It returns to its maximum at the end of the second stimulus. }
\label{fig:pred-schematic}
\end{figure}

As shown in Fig.~\ref{fig:pred-schematic}, we consider a signaling system based on signal strength as a threshold to assign meaning to the signal (i.e., ground the signal). Depending on the informational content of the signal (i.e., token value), the signal could carry information regarding prediction of the occurrence of an event. For example, in an anticipatory prediction (Fig.~\ref{fig:pred-schematic}a), if the value of a prediction is larger than a threshold, an event is predicted to occur in the near future. In a time-to-event style of question prediction (Fig.~\ref{fig:pred-schematic}b), if the value of prediction is smaller than a threshold, an event is predicted to occur in the near future. As can be seen in the signalling system used in the two cases in Fig.~\ref{fig:pred-schematic}, the grounding rule used to ground the tokens are different, however in both cases token value of 1 means the next event will occur "soon", and a token value of 0 means the next event is not predicted to be "soon".

What we call Pavlovian signalling has been deployed in the literature for agent-agent interaction, but to our knowledge not clearly defined or explored in depth. Examples from the literature mapped learned predictions to scalar vibratory tokens for a human receiver~\cite{edwards2016}, mapped learned predictions about viable foraging locations to audio feedback~\cite{pilarski2019}, and mapped learned predictions of robot sensor activation to signalling tokens via a fixed threshold that was grounded in the values of real-world motor sensors~\cite{parker2019} . In these examples, a hard-coded mapping was created from learned predictions to emitted tokens, where tokens were defined using parameters of the predictive questions being learned by the machine and the sensory threshold values used to define the mapping process. This paper provides a foundation for understanding the impact of such prediction and mapping choices on the efficacy of resulting agent-agent interactions. 

\section{Methods: Predictions, representations, and tokens}

We now turn to the specific computational framing we use for predictions and signals in this work. We herein model an agent's predictions about the world using General Value Functions (GVFs) which are value functions applied to non-reward based targets \cite{sutton2011}. A GVF formally specifies a predictive question, which can be understood informally as: {\em what will be the total accumulation of some signal of interest, if I follow some policy until termination?} A GVF is a value function where the target is the discounted sum of some cumulant $C_{t+1} \in \mathbb{R}$, that would be observed if the agent followed policy $\pi(A_t|S_t) \doteq Pr(A_t = a| S_t = s)$---a {\em what would happen if} form of question. The elements of the sum are weighted by earlier discounts $\gamma_t \in \mathbb{R}$. The discount becomes zero if a termination event occurs, and is typically less than one otherwise corresponding to the horizon of the predictive question. Taken together, we can  specify a predictive question by defined $C, \gamma$, and $\pi$. We first define the future return,
\begin{equation}G_t \equiv \sum_{k=0}^\infty \Bigg(\prod_{j=1}^k\gamma_{t+j}\Bigg)C_{t+k+1},\label{eq:gvf}\end{equation}
where the question is then defined as
\begin{equation}v(s) \equiv \mathbb{E}_\pi \{G_t | S_t = s\}.\label{eq:gvf_question}\end{equation}

The agent must learn answers to the GVF question to obtain knowledge of the world; that is, approximate $v$ from data. Given a batch of data, we could compute the right-hand side of Equation \ref{eq:gvf} directly. In practice, the agent will observe a stream of states, actions, cumulants, and terminations as it interacts with the world. In this online setting, we can approximate $v$ in each state with a parametric function updated via temporal difference learning. Let $x \in \mathbb{R}^d$ be features summarizing the current state $x_t \equiv x(S_t)$, perhaps a state aggregation or a collection of radial basis function outputs. We define the prediction to be $V_t \equiv w_t ^\intercal x_t$, where $w_t \in \mathbb{R}^d$ and $V_t \approx v(S_t)$. Although more complex methods are possible, we follow prior work~\cite{sutton2011,modayil2014multi} and use the TD($\lambda$) algorithm to update $w_t$ on each timestep:  
\begin{align*}
e_t &\gets e_{t-1} + x_t\\
\delta_t &\gets C_{t+1} + \gamma(x_{t+1}) w_t ^\intercal x_{t+1} -  w_t ^\intercal x_t\\
w_{t+1} &\gets w_t + \alpha \delta_t e_t\\ 
e_{t} &\gets \gamma(x_{t+1}) \lambda e_{t},
\end{align*}

where $\alpha$ is a scalar learning rate parameter and $e \in \mathbb{R}^d$ is exponentially decaying memory of previous feature activations. This approach has been shown to learn accurate approximations of GVF questions in a variety of settings and is computationally frugal---requiring computation and memory linear in the number of features $d$---and is thus ideal for our problem setting of interest.  

For this work we consider the two specific types of predictive GVF questions shown in Fig. \ref{fig:pred-schematic}: predictions about the onset of an impending signal or stimulus in terms of the expected future accumulation of that signal (a {\em rising} prediction about a future event), and a prediction of the expected time remaining until a future signal or stimulus pattern (a {\em falling} prediction or learned countdown timer until an event will occur). Both of these predictive questions of interest can then be specified in terms of the three GVF question parameters---the signal of interest is specified as the cumulant $C$, the timescale of interest as specified by $\gamma$, and the policy $\pi$ of interest.

First, for the case of a discounted accumulation of an observed stimulus, we can identify GVF question parameters in what we refer to informally as an {\bf \em accumulation question}. The cumulant takes the value of a specific stimulus, and the gamma-discounted sum of the cumulants gives lower emphasis on stimuli farther in the future: $C_t = 1.0$ if stimulus is present else $0.0$, $\gamma_t = 0.9$, and $\pi = \text{on policy}$.

Second, we consider a learned prediction of the expected number of steps until an observed stimulus, which we will refer to informally as a {\bf \em countdown question}. A constant cumulant of 1.0 is used on every step, with question termination occurring at the onset of a stimulus as follows: $C_t = 1.0$, $\gamma_t = 0.0$ if stimulus is present else $1.0$, and $\pi = \text{on policy}$.

With these questions, we can directly implement the process for Pavlovian signalling depicted in Fig. \ref{fig:pred-schematic} wherein predictions are mapped to Boolean tokens according to fixed thresholds. For example, in fixed-time scale prediction, if the value of the prediction rises to exceed a fixed, scalar threshold, a token of 1 is emitted. If the prediction is less than or equal to the threshold, a token of 0 is emitted. Similarly, in the countdown prediction, if the value of the prediction decreases to be equal to or less than a threshold, a token of 1 is emitted (with the token of 0 otherwise).

\subsection{Temporal Representations}
\label{sec:representation-defs}

Multi-agent temporal decision-making tasks are the key focus of the present work. Making predictions about events that unfold over time requires a sufficiently informative representation, $x_t$. We therefore implement a subset of biologically and computationally motivated temporal representations as a basis for GVF learning (sampled from present literature \citep{buonomano2017,eichenbaum2014,eichenbaum2017,paton2018, hochreiter1997, Neil2016, Tallec2017, koutnik2014}), such that we can understand the sensitivity (or lack thereof) of prediction learning speed and agent performance to temporal representation. 
In biological agents, distinctions have been made between {\em population clocks} and  {\em ramping models} \citep{tsao2018,paton2018,eichenbaum2014} (c.f., pacemaker-accumulator models \cite{paton2018}, following on seminal work on scalar-expectancy theory  \cite{gibbon1977}).
Population clocks are considered a chain or collection of cells that sequentially fire, either in response to outside inputs or in an oscillatory fashion \citep{paton2018}. In ramping models, time is represented in changes to the tonic firing rate of different cell populations \cite{paton2018}. Trace-conditioning \cite{pavlov1927}, akin to ramping model use, was also investigated in a machine learning context by Rafiee et al. \cite{rafiee2021}.
 
The temporal representations used in this study draw directly on the mechanics of population clocks, ramping models, and trace conditioning as surveyed above. Our representations all take as input a bit indicating the presence of a specific stimulus of interest---a {\bf \em presence representation (PR)} \cite{rafiee2021}---and output a one-hot vector of some length that indicates a position in time since the last stimulus. The length of the vector and the rate of movement of the active bit within it vary with each treatment. The stimulus presence bit and the time representation vector are then concatenated and used as the state representation for a GVF. The simplest temporal representation included in the study, the {\bf \em bias unit representation}, was implemented as a vector with one constant feature, or equivalently, a one-hot vector of length one. Following on the general form of population clocks from the animal brain, our implementation of the {\bf \em bit cascade (BC) representation} extended this one-hot vector to length $n$, where $n$ was longer than the maximum time anticipated between occurrences of the specific stimulus. The active bit advanced by one position on each timestep where the stimulus was absent, and was reset to the first position whenever the stimulus was present. Finally, and most closely aligning to biological ramping models, the {\bf \em tile-coded trace (TCT) representation} was an extension of the bit cascade with advancement rates of the exponential form \(e^{-at}\), with \(a < 1.0\) representing the exponential decay constant, and where $t$ is the number of timesteps since the last reset of the active element to the beginning of the vector. As with the bit cascade, the active bit was reset when the stimulus was present. Thus, each position in the vector represented an increasingly long span of time as the bit advanced. See the left column of Fig.~\ref{fig:nexting-short-isi-comp} for an example of these representations of time.

\section{The Frost Hollow Environment}

\begin{figure*}[!t]
\centering
(a)\ \  \includegraphics[height=1.5in]{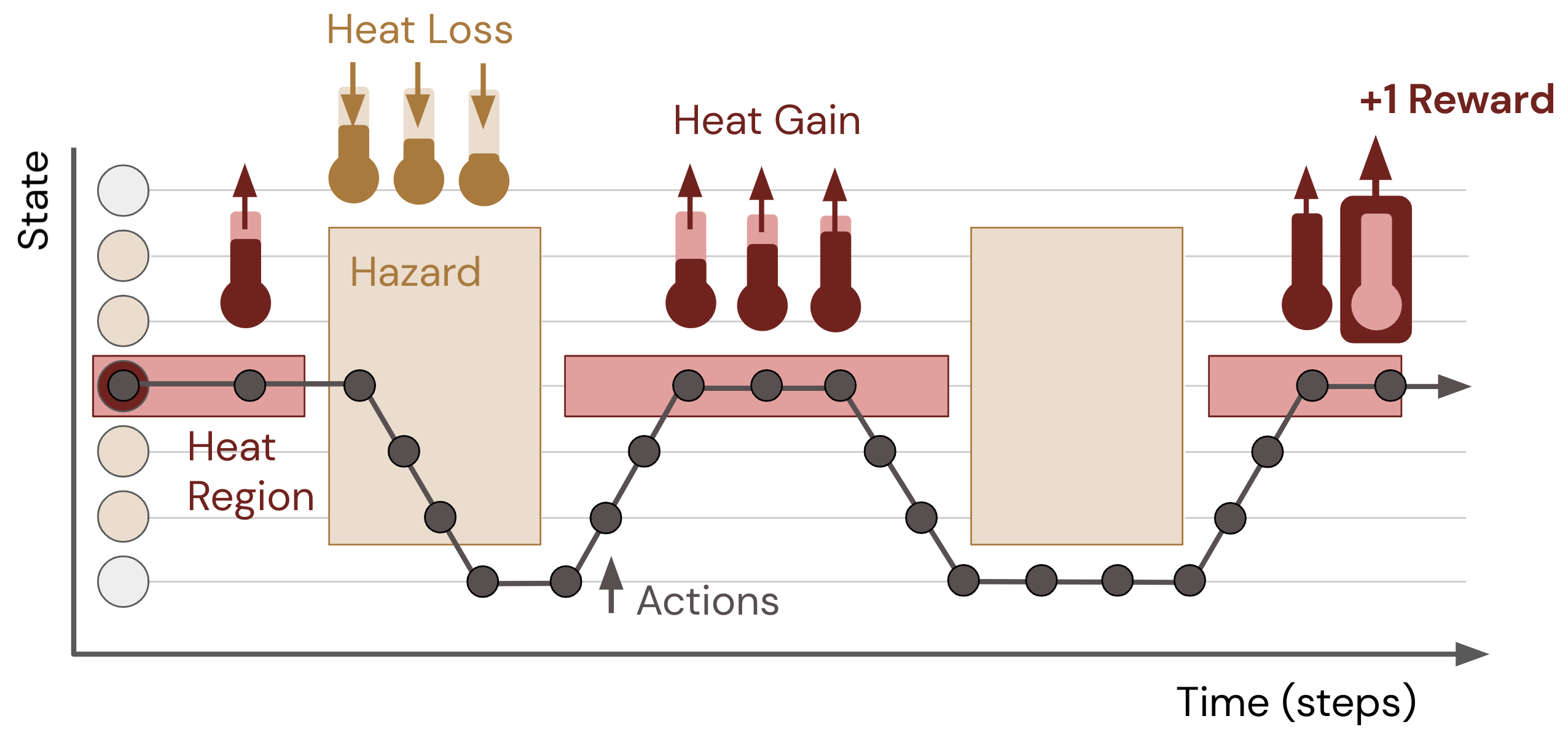} \hfil
(b)\ \ \includegraphics[height=1.5in]{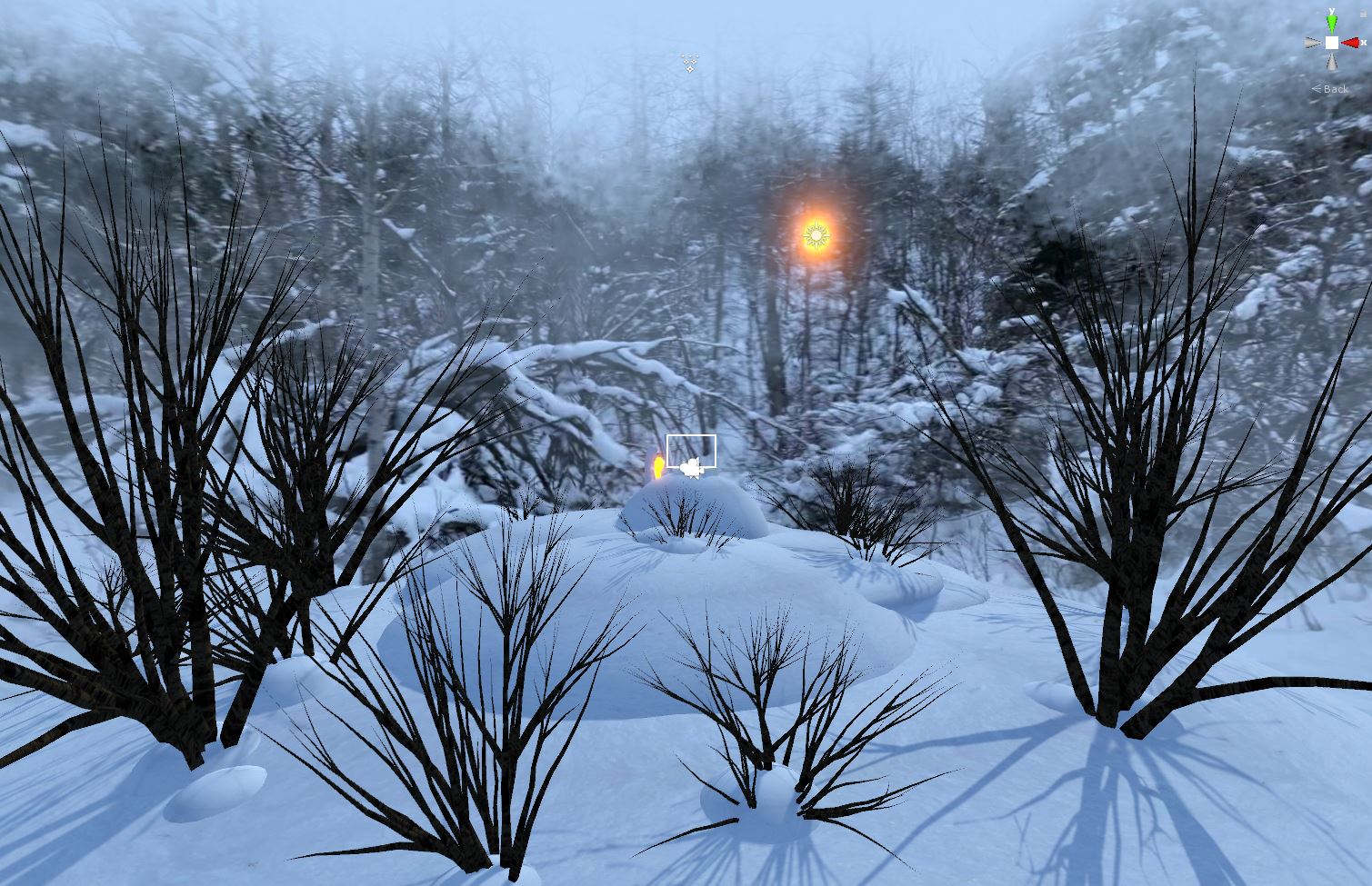}
\caption{Abstract (a) and virtual reality (b) variants of the Frost Hollow domain, and one example trajectory showing how an agent might interact with the abstract version over time. The goal is to maximize the amount of \textit{reward} a player earns.}
\Description{Two panels show the abstract and virtual reality environments side by side. The temporal sequence of an agent taking actions and moving in the abstract environment is shown in the first panel. A specific trajectory is plotted that shows the agent gaining heat by standing in the heat region, and being hit by a hazard while attempting to move to safety. The agent returns to the heat region, avoids the next hazard, and returns again to the heat region to gain additional heat and obtain a reward. The second panel shows a rendering from the virtual reality Frost Hollow environment. }
\label{fig:abstract-frosthollow}
\end{figure*}

Researchers in machine learning have used simulation problems inspired by animal learning to better understand the capabilities of artificial agents. Drawing on this rich history of animal and machine learning experimentation, and in particular recent work by Rafiee et al. \cite{rafiee2021} on trace conditioning and a suite of problems inspired by experiments in animal learning, we now introduce a domain for our study called {\em the Frost Hollow}. 

The Frost Hollow environment is a partially observable domain to evaluate our agents' ability to predict and generate signals relating to events that unfold over time---in this case, collecting heat from sunlight and avoiding heat loss due to the hazardous winter wind. Figure \ref{fig:abstract-frosthollow} presents this domain and defines key elements of the environment, namely reward, heat region, heat points, heat capacity, the wind hazard and the hazard region.
From this starting point, we created two environment variants, an abstract linear-walk domain for agent-agent interaction and a virtual reality (VR) variant for human-agent interaction (shown in Fig.~\ref{fig:abstract-frosthollow}). In each episode of either variant, the player's goal is to collect a specific amount of heat to gain one point of reward, and to do so repeatedly to maximize their total episodic reward. Within each episode, a cold wind blows somewhat regularly according to a set of parameters, and being exposed to the wind causes the player to lose any accumulated heat. The player can avoid the wind by moving to the edge of the map before it blows. Thus, the core challenge of the environment is to predict hazard events over time: using timing features about the past (e.g., how long since the previous wind hazard) to make useful predictions about the future (e.g., how long until the next hazard), and use those predictions to guide behavior. 

The {\bf Abstract environment} is a chain of seven locations, as shown on the y-axis in Fig.~\ref{fig:abstract-frosthollow}a. On each timestep the environment provides an observation containing a one-hot vector indicating agent location, a boolean indicating whether or not the hazard is active, and the agent's accumulated heat as a scalar in [0, heat capacity]. The farthest locations at each end of the chain are not affected by the hazard. The action space is an integer in [-1, 0, 1], to move down, stand still, or move up through the map locations. Heat is gained at a rate of $0.5$ per timestep in the heat region, automatically converted into 1 point of reward when the target heat capacity of 6 is reached, and set to 0 if the agent is exposed to the hazard.

The {\bf Virtual Reality environment} is presented to a human player using the Valve Index VR headset, which displays a 1440x1600 pixel display at 120 frames/second. The player uses their body to walk through a 3 meter by 2 meter area, divided with concentric circles into a 0.165 meter radius heat region in the center, surrounded by a 1 meter radius hazard region. The area outside of the hazard region, at the edges of the map, are a safe region where they are protected from the hazard. The player visually observes their position, the presence of the hazard, and amount of heat. The player also receives input via controller vibration indicating the presence of the hazard. Pavlovian signalling to the player was also effected by vibrating a second controller. The player's action space is to walk and move their headset through space, the position of which is mapped into the virtual environment to determine which region the player is in. Heat is gained at a rate of $0.1875$ per second in the heat region, converted into reward in the heat region when the player has accumulated 5 heat points by raising a controller over their head, and set to 0 if the player is exposed to the hazard.

\begin{figure}[!th]
\centering
{\bf REPRESENTATION \hfil PREDICTIONS}\\
\includegraphics[width=1.5in]{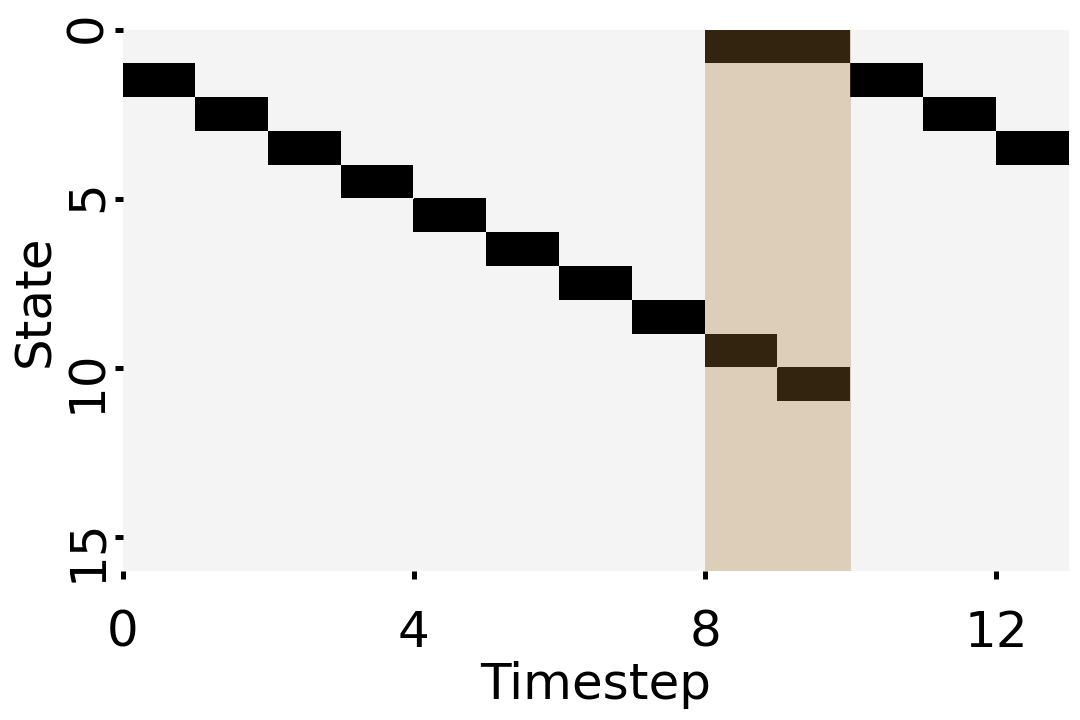}
\includegraphics[width=1.5in]{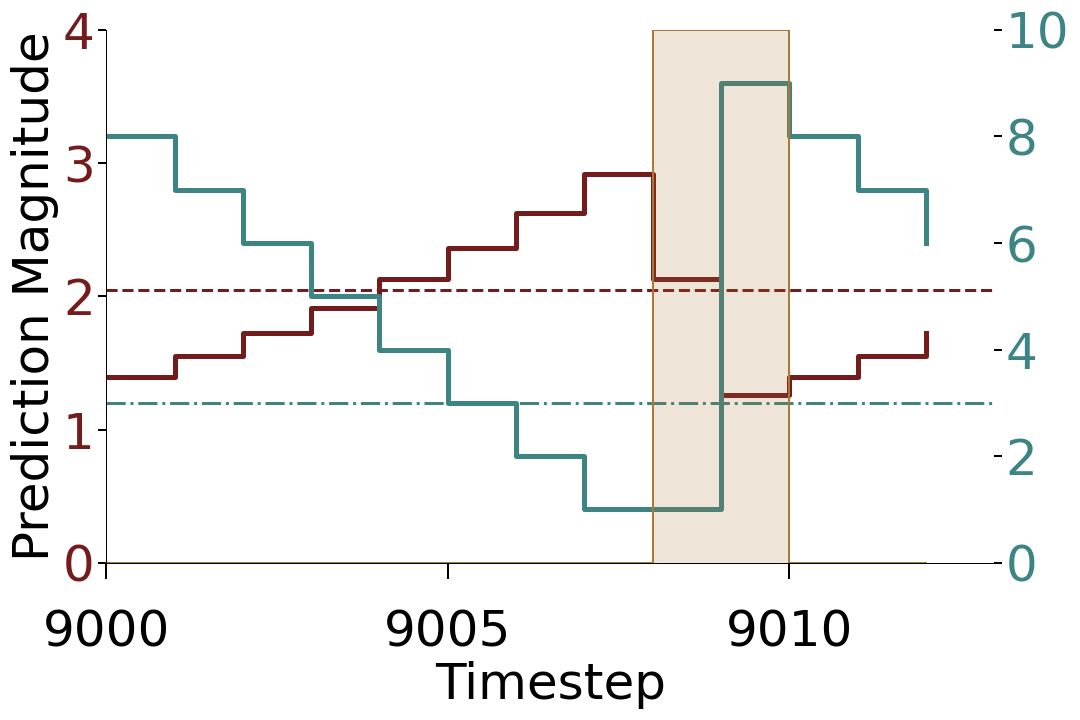}\\
(a) { Bit Cascade + PR}\\
\ \\
\includegraphics[width=1.5in]{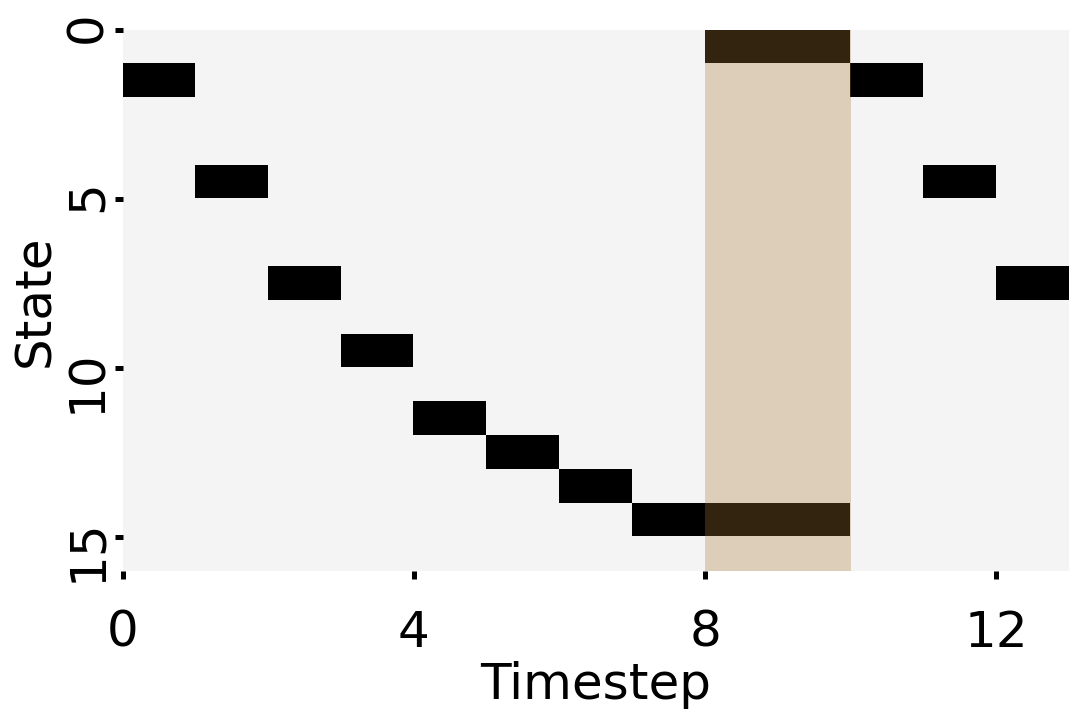}
\includegraphics[width=1.5in]{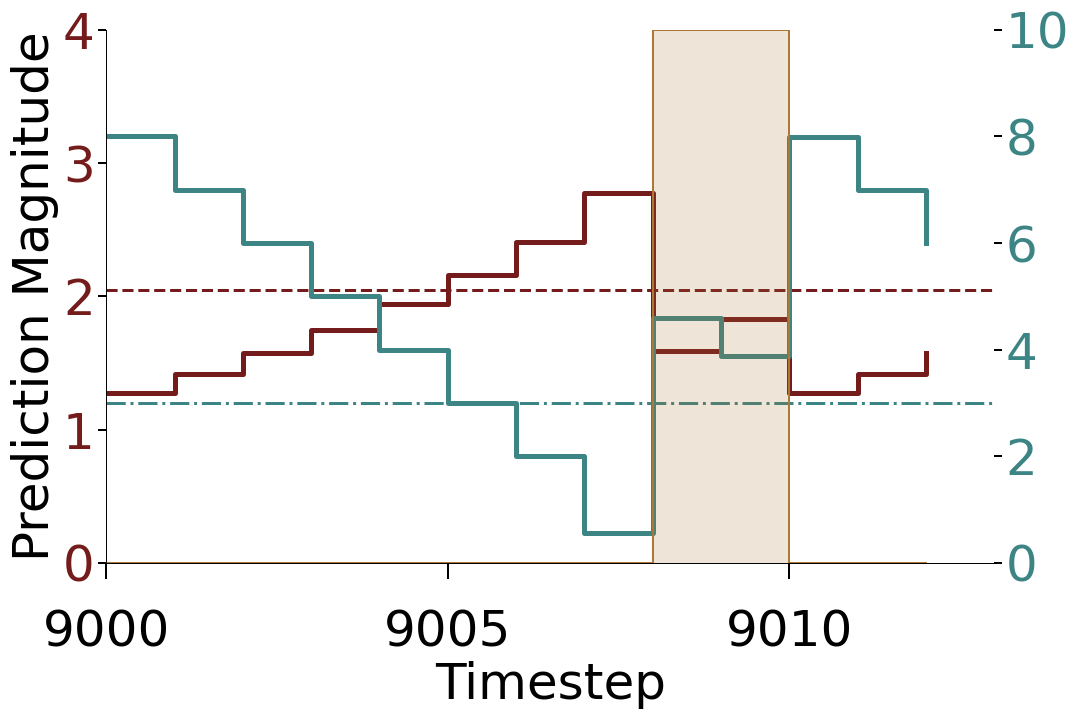}\\
(b) { TCT + PR} (decay = $e^{-0.3t}$)\\
\ \\
\includegraphics[width=1.5in]{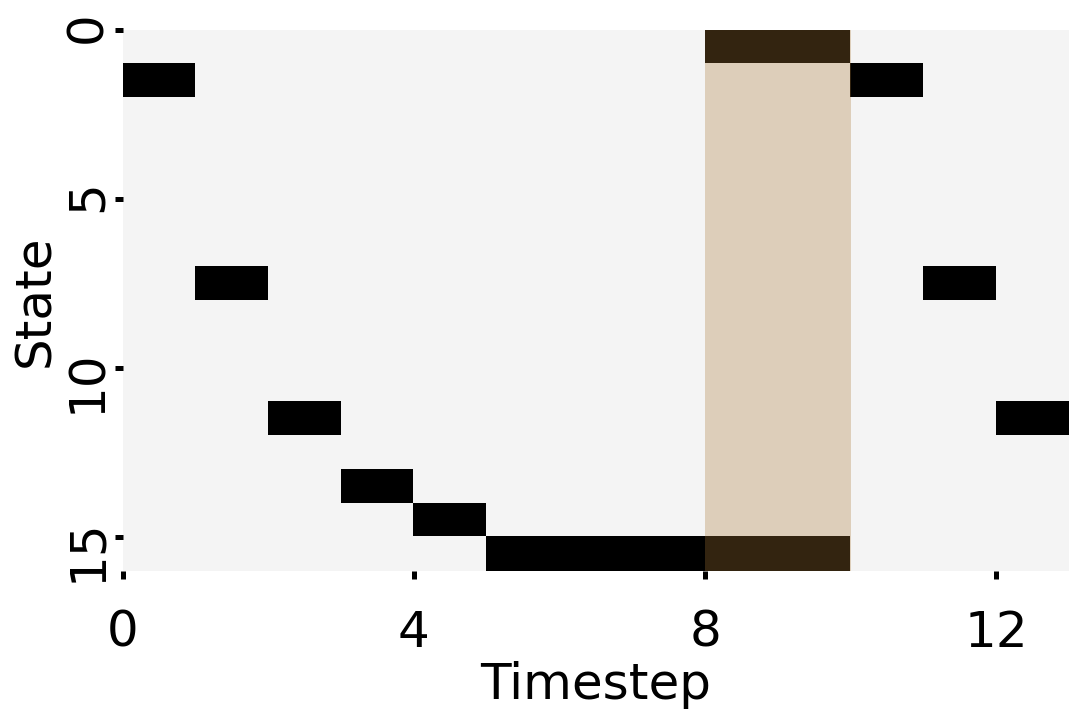}
\includegraphics[width=1.5in]{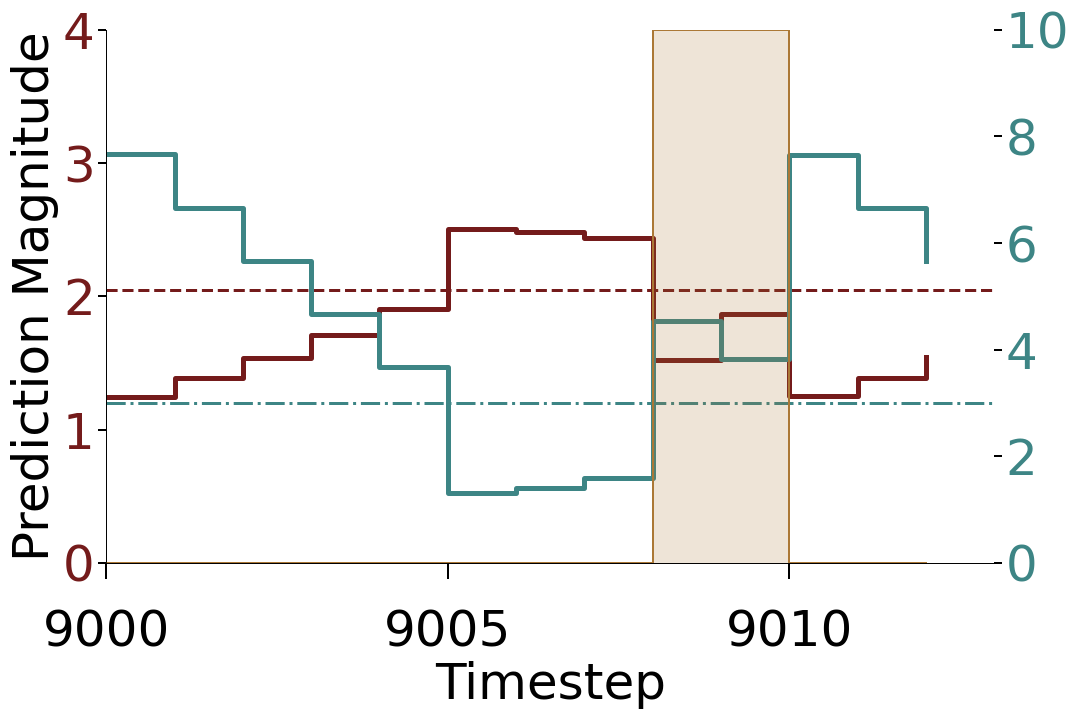}\\
(c) { TCT + PR} (decay = $e^{-0.6t}$)\\
\ \\
\includegraphics[width=1.5in]{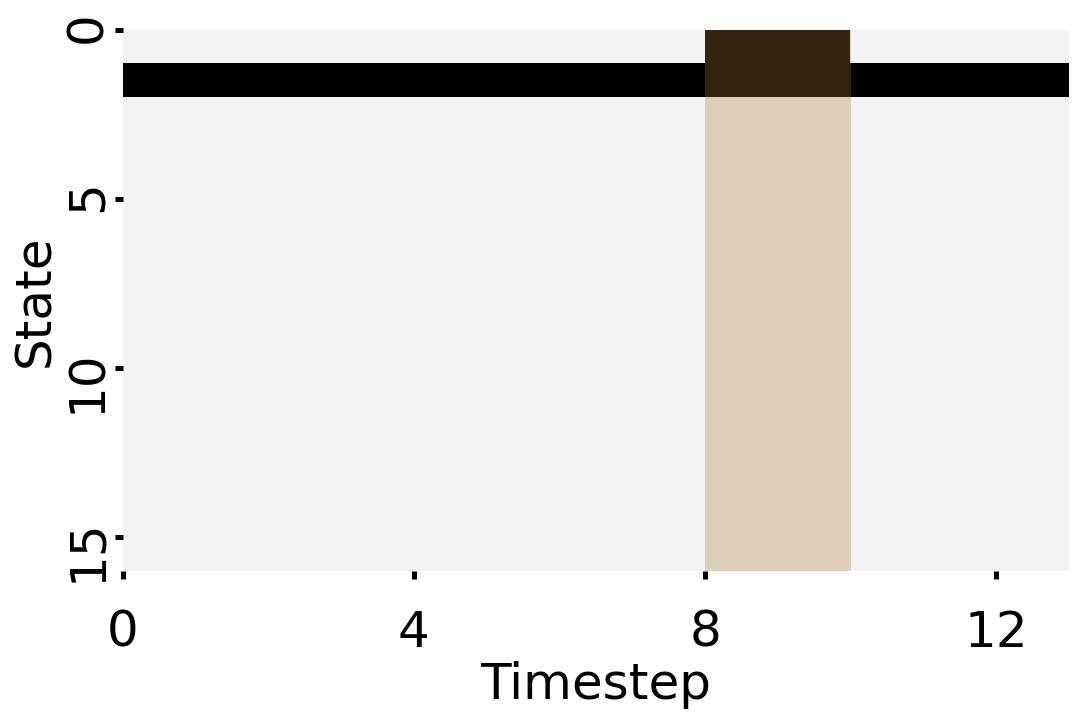}
\includegraphics[width=1.5in]{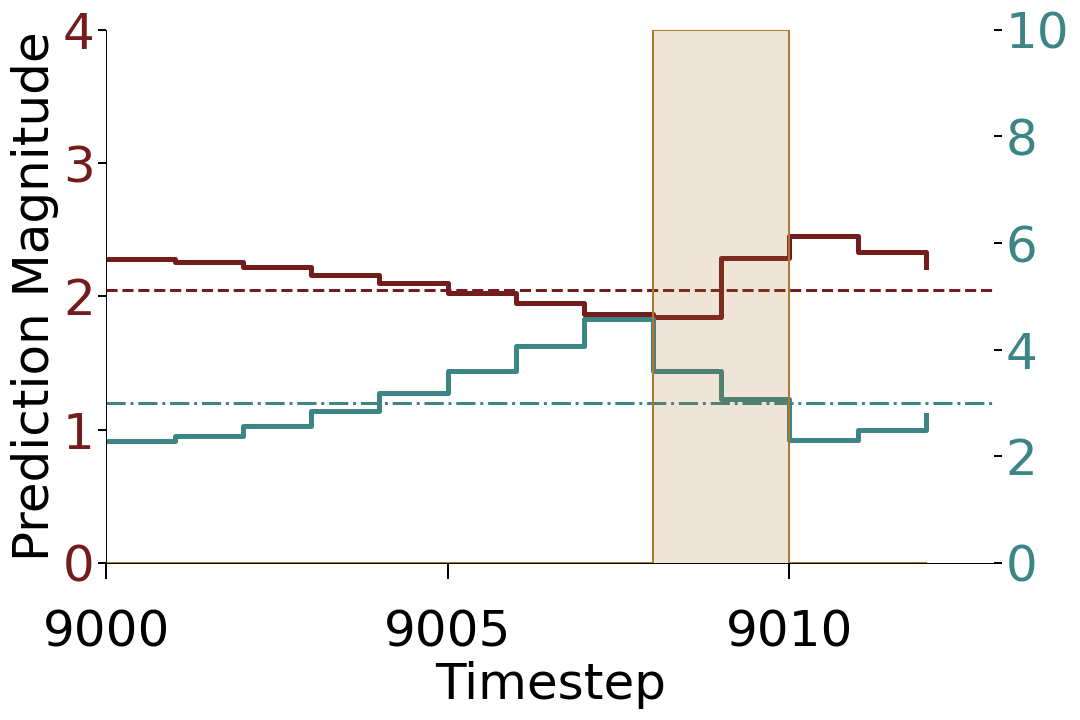}\\
(d) { Bias Unit + PR}\\
\caption{Examples of GVF learning in the Frost Hollow domain for each of the fixed conditions.  The left column shows recorded bit progressions for each representation over spans of single ISIs of equal length, including the presence representation (black square in the top row of each plot) and the temporal representation (black squares in rows 1+). The right column shows an example of prediction learning for both the accumulation (red trace) and countdown (blue trace) GVF questions with four different representations (hazards shown in yellow). The threshold levels for Pavlovian signalling are shown as horizontal dashed lines. Note that with a Bias Unit time representation, which cannot distinguish states temporally, the weight updates move such that the predictions are backwards: a countdown GVF trends up until a stimulus, and an accumulation trends down.}
\Description{In the left column the presence bit and the temporal representation are shown unfolding over time. In the right column, the predictions made by an accumulating and a countdown co-agent are dual-plotted. The tokenization threshold for each question is indicated with a horizontal line. 

Each representation permits tokens that change between 0 and 1 appropriately, but temporal aliasing changes prediction magnitude, and consequently the times at which tokens switch from 0 to 1 and back.}
\label{fig:nexting-short-isi-comp}
\end{figure}

The {\bf hazard experimental conditions} are described by two key variables: the \textit{inter-stimulus interval} (ISI) which is the time between hazards, and the \textit{stimulus length} which is the duration of the hazard; see Figure~\ref{fig:nexting-short-isi-comp} for an example. With this framework, we considered three types of hazard: {\bf \em fixed} ISI, {\bf \em random} ISI where the inactive period of the ISI was varied randomly between upper and lower bounds, and a {\bf \em drift} ISI wherein the hazard has a fixed stimulus length and ISI interval (e.g., 2 and 8 in the abstract environment, as in the fixed condition), but the ISI varies permanently by $\pm n$, within a minimum and maximum range, before each hazard (e.g., $\pm 1$ within the bounds [6, 11] in the abstract environment). 

\section{Control Learning in Frost Hollow with Pavlovian Signalling}
\label{sec:control-experiments}

\begin{figure}[!t]
\centering
\includegraphics[width=0.9\linewidth]{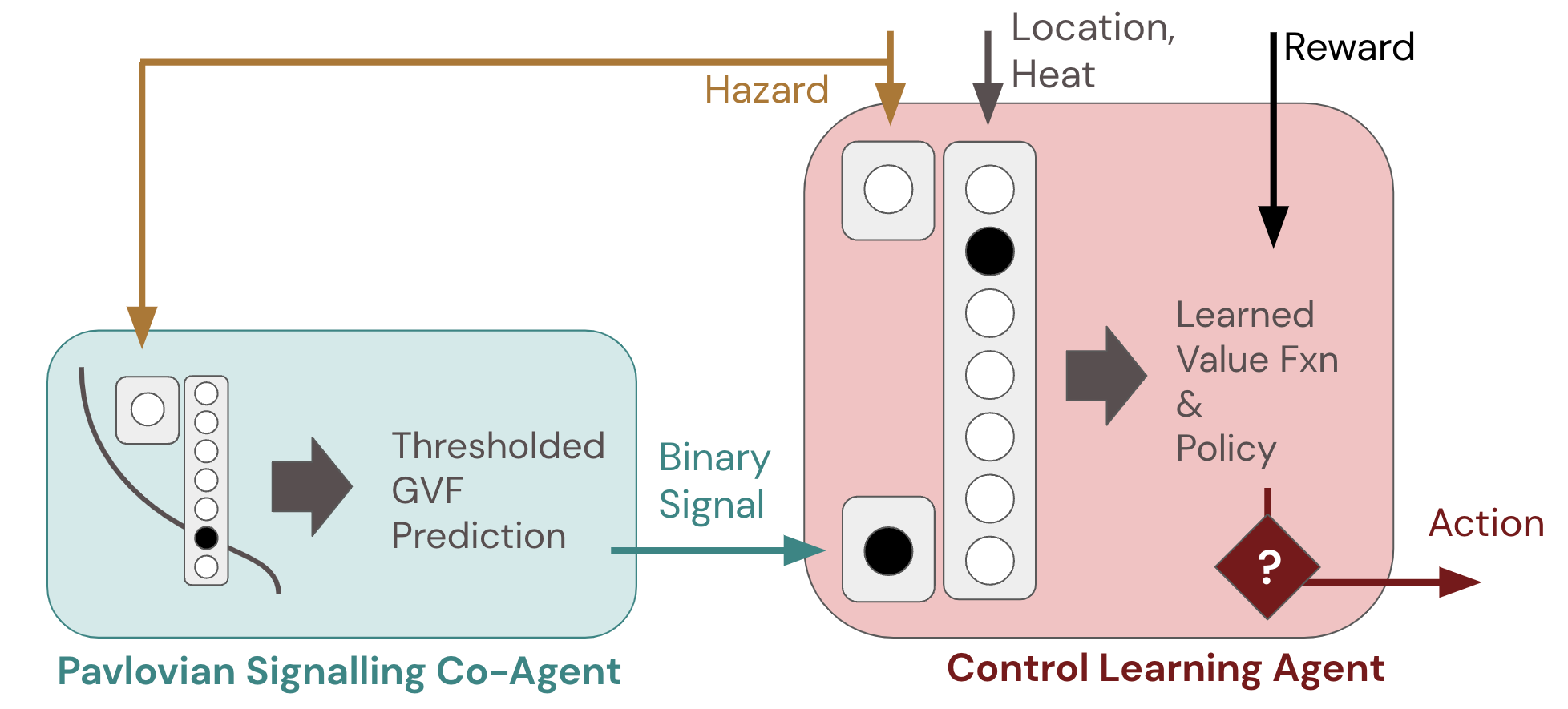}\\
\caption{Pavlovian signalling co-agent that learns to predict a hazard and passes tokens to a control learning agent.} 
\Description{The Pavlovian Signalling Co-Agent and the Learning Agent are displayed side by side. Hazard information is shown going to both the agent and the co-agent. Only the agent receives location and reward information. The co-agent concatenates the hazard information with the state of an internal temporal representation, and uses it to generate a binary signal which is the only temporal information available to the learning agent. The Learning Agent acts in the environment.}
\label{fig:agent-agent-schematic}
\end{figure}

As shown in Fig. \ref{fig:nexting-short-isi-comp}, there were key differences in how temporal representations were able to support the prediction learning with respect to phenomena unfolding in time, and the way these predictions might be turned into tokens that could be used by a second agent or a discrete decision-making unit within a single agent. For this section we therefore specifically consider the case where we have two interacting learning machines that share one avatar in the Frost Hollow domain and that must interact so as to collect reward. While it is natural to think of these two learning machines as two tightly coupled parts of a single larger learning machine, here we depict them as concrete independent learning machines. We do this so as to be able to clearly describe the comparisons we make between different conditions without creating assumptions about other parts of a single-agent composite architecture, and also to align with the human-machine virtual reality experiments presented below.
 
We denote the two learning machines under consideration as the {\em agent} and the {\em co-agent}. The agent is responsible for policy learning so as to solve the Frost Hollow control challenge, while the co-agent is responsible for prediction learning so as to provide accurate and relevant forecasts about the Frost Hollow hazard stimulus. As shown in Fig.~\ref{fig:agent-agent-schematic}, the only state input to the co-agent in our experiments is the presence representation, a single bit that indicates the presence or absence of the hazard on a given timestep. All other state information for the co-agent is derived from its internal temporal representations. The output of the co-agent is a single bit: a token that takes either the value of 1 or 0, indicating its prediction of an imminent hazard. In contrast, the agent is a control learner that takes as input the reward from the environment and, as a tabular representation, the token from the co-agent, the presence representation for the hazard, and its location and heat (also shown in Fig.~\ref{fig:agent-agent-schematic}). 

In our experiments we will use two types of co-agents paired with a learning agent. The {\bf \em Pavlovian signalling co-agent} uses one of the three types of temporal representations described in Section~\ref{sec:representation-defs} coupled with an accumulation GVF or a countdown GVF. The resulting prediction is compared against a fixed threshold to produce the signal sent to the agent. The {\bf \em oracle co-agent} is used for comparison as an artificial upper bound on the quality of information we could expect from a co-agent. Instead of learning, it uses a fixed policy that observes hidden information from the environment about exactly when the next hazard will arrive, and emits a token of 1 whenever it must start moving to safety in time to avoid the hazard. The {\bf \em control learning agent} uses on-policy reinforcement learning with the state representation described above, and learns through trial and error how to gain reward.

One final consideration for Pavlovian signalling co-agents is the value of the fixed threshold which is compared to predictions to produce tokens. This threshold determines when a stimulus is predicted to be sufficiently soon that the control learning agent must be notified, and thus accounts for factors such as the lead time needed for the control learning agent to react, inaccuracy in the prediction, and stochasticity in the environment. In our experiments we selected the threshold values based on the advance notice needed for an agent in the abstract domain to avoid hazards and thus accumulate heat for reward, as can be arrived at empirically or from the idealized return of Eq. \ref{eq:gvf}: $\tau=2.05$ for the accumulation GVF and $\tau=3.0$ for the countdown GVF---e.g., for a countdown GVF specified by $C$ and $\gamma$, when the expected steps until the hazard $V_t=3.0$ an agent would have the needed 3 steps to move from the heat region to safety.

The control learning agent follows the standard Expected Sarsa($\lambda$) learning algorithm 
\cite{sutton2018}, chosen to minimize complexity on the agent side and focus on guiding our understanding of paired agent/co-agent learning and the role of temporal abstractions in learning dynamics. For reference here, Expected Sarsa($\lambda$) calculates its temporal-difference error $\delta_t$ via a summation over all actions weighted by their probability under the current policy $\pi$ for the future state $x(S_{t+1},A_{t+1})$, as combined with the reward $R_{t+1}$ and the action values for the current state $x(S_{t},A_{t})$, and uses this error and eligibility traces $e_t$ to updates the weights $w_t$ associated with its action values as follows:
\begin{align*}
e_t &\gets e_{t-1} + x(S_{t},A_{t})\\
\delta_t &\gets R_{t+1} + \gamma \sum_a [\pi(a|S_{t+1}) w_t ^\intercal x(S_{t+1},A_{t+1})]\\
w_{t+1} &\gets w_t + \alpha \delta_t e_t\\ 
e_{t} &\gets \gamma \lambda e_{t}
\end{align*}
Here $\alpha$ is the step size for learning weights, $\lambda$ is the eligibility trace decay rate, and $\gamma$ is the discounting rate applied to future actions values; action values used in action selection for a given state $S_t$ and action $A_t$ are approximated via the linear combination $Q(S_t,A_t) = w_t ^\intercal x(S_t,A_t)$, where action values were optimistically initialized and were selected on each step according to epsilon-greedy action selection. We examined the performance of co-agents with countdown and fixed-gamma GVF questions with bias, bit cascade, and tile-coded trace representations with GVF learning rates $\alpha_{gvf} \in \{0.01, 0.1\}$ to study fast and slow tracking of non-stationarity in the environment. Control learning algorithm parameters were determined via empirical sweeps, with results below shown for the best-case values of $\alpha=0.01$, with exploration via an $\epsilon$-greedy exploration policy, $\epsilon \in \{0.01, 0.1\}$, and optimistic initialization with weights initialized to 1.0. Token generation with respect to prediction magnitude can be seen in the relationship between solid and dashed lines in Fig. \ref{fig:nexting-short-isi-comp}.

\section{Abstract Environment Results}

\begin{figure}[!t]
\centering
{\bf Accumulation GVF} \hfil {\bf Countdown GVF}\\
\includegraphics[width=1.6in]{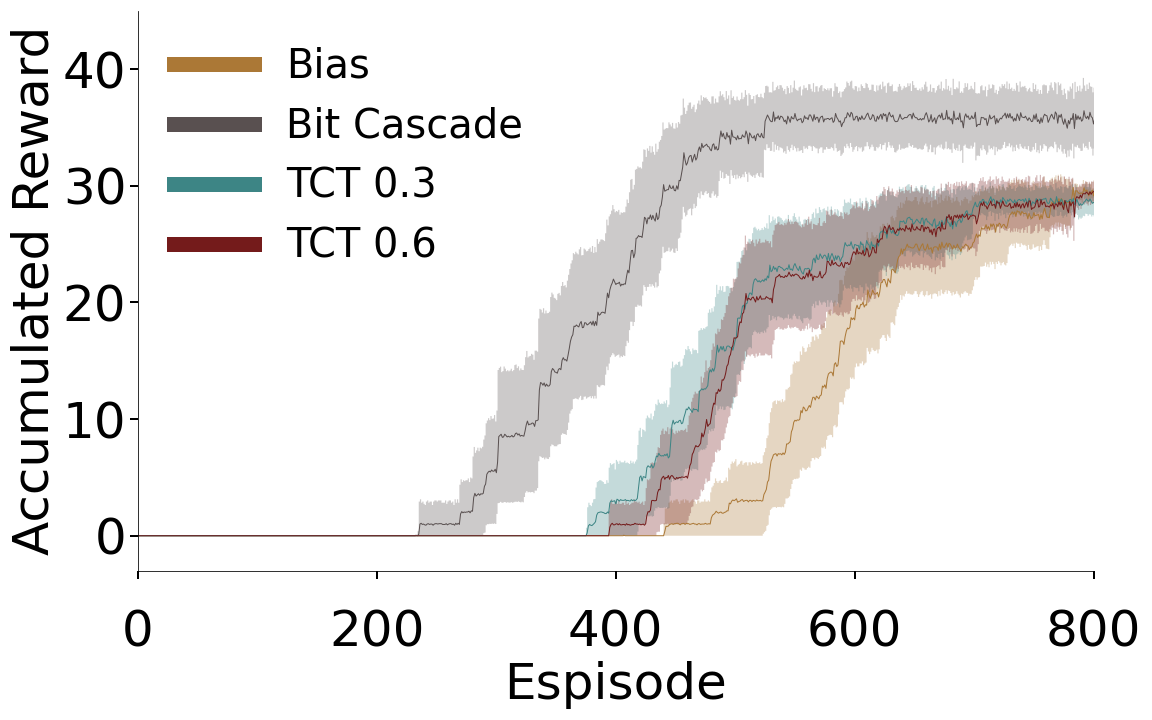}
\includegraphics[width=1.6in]{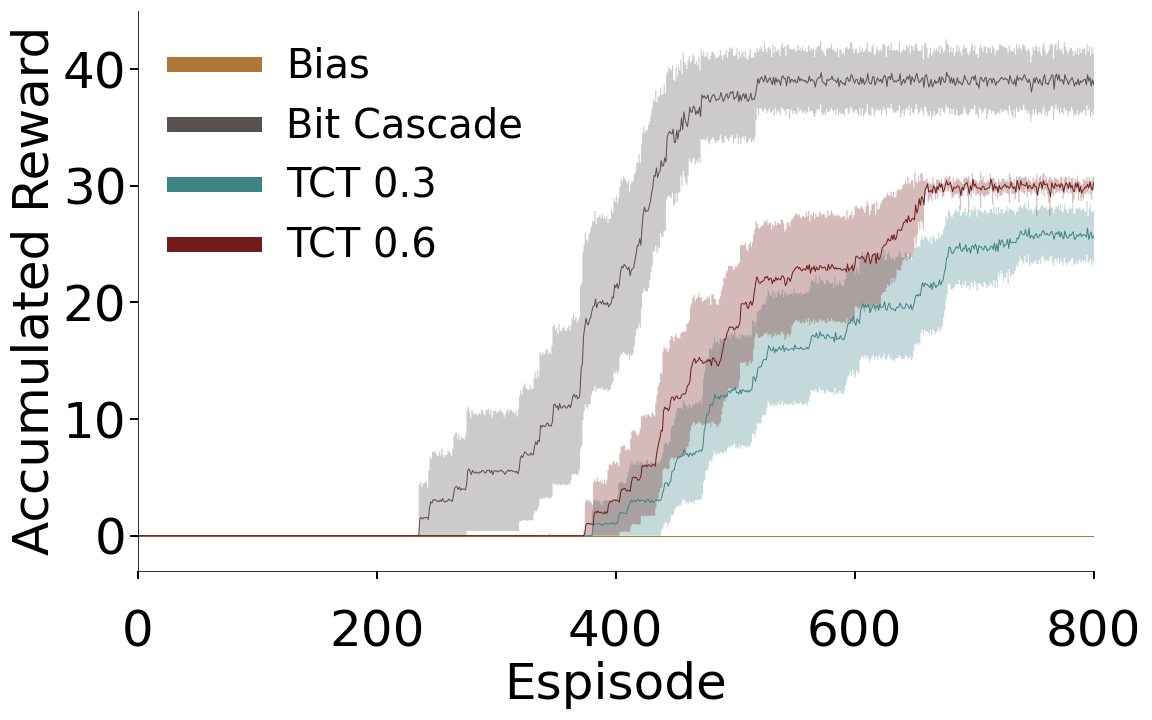}
(a) { FIXED} environmental condition\\
\ \\
\includegraphics[width=1.6in]{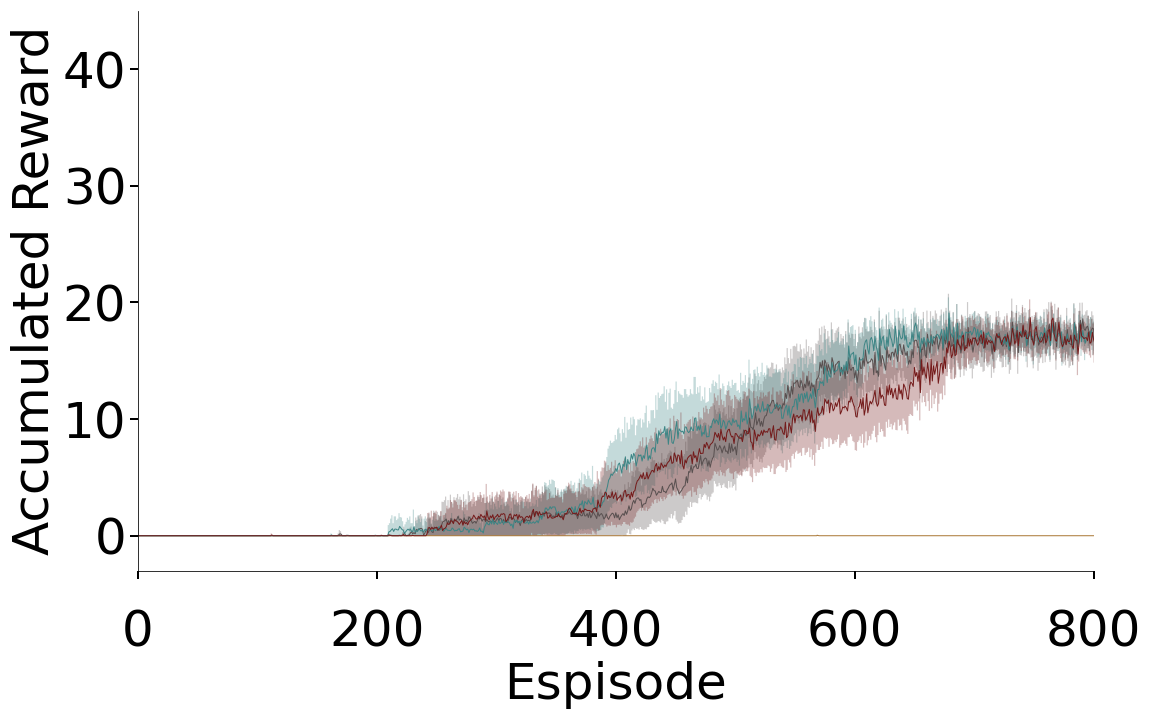}
\includegraphics[width=1.6in]{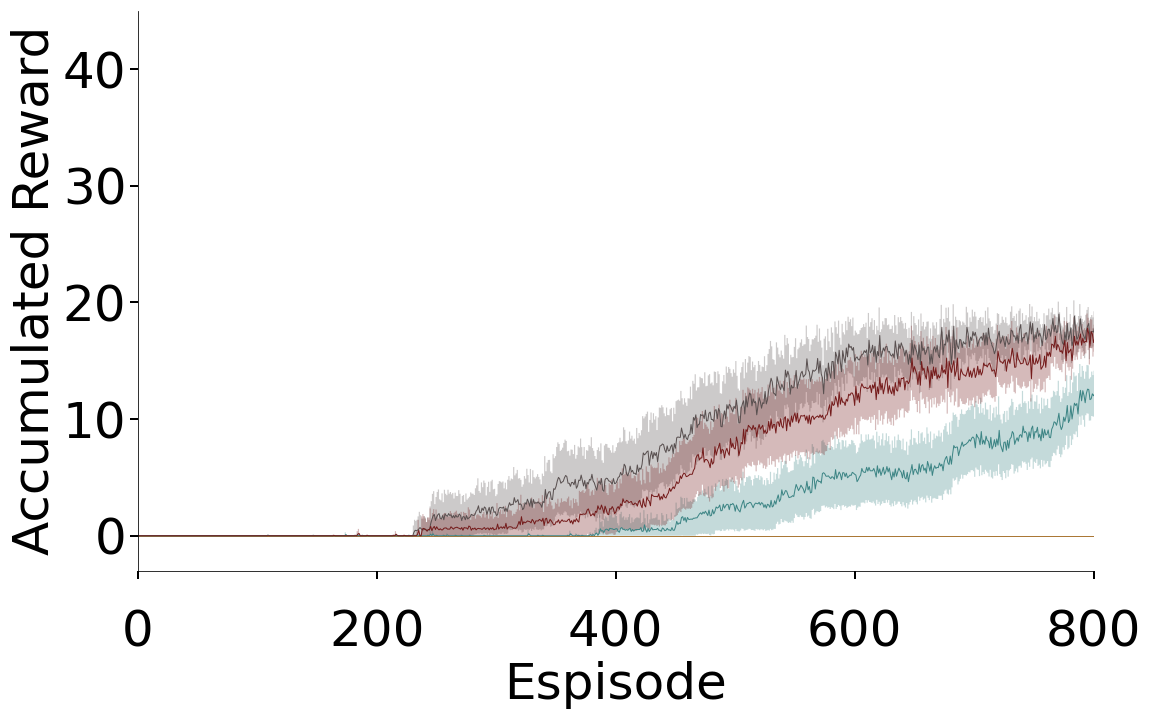}
(b) { RANDOM} environmental condition\\
\ \\
\includegraphics[width=1.6in]{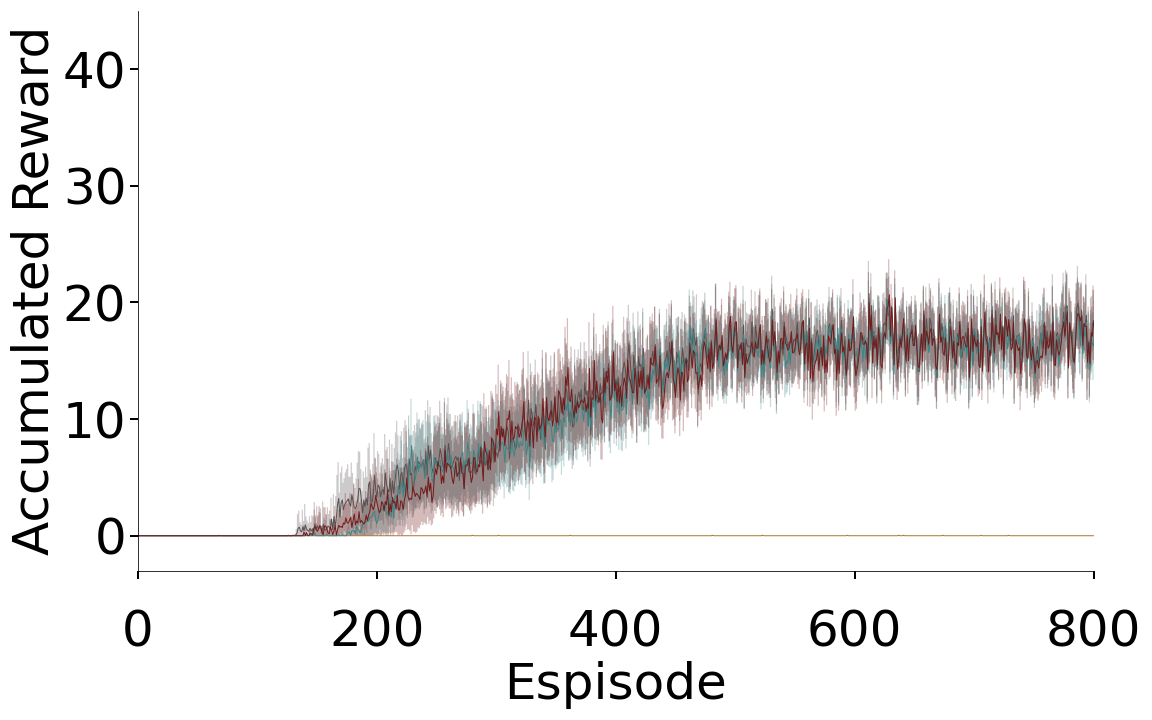}
\includegraphics[width=1.6in]{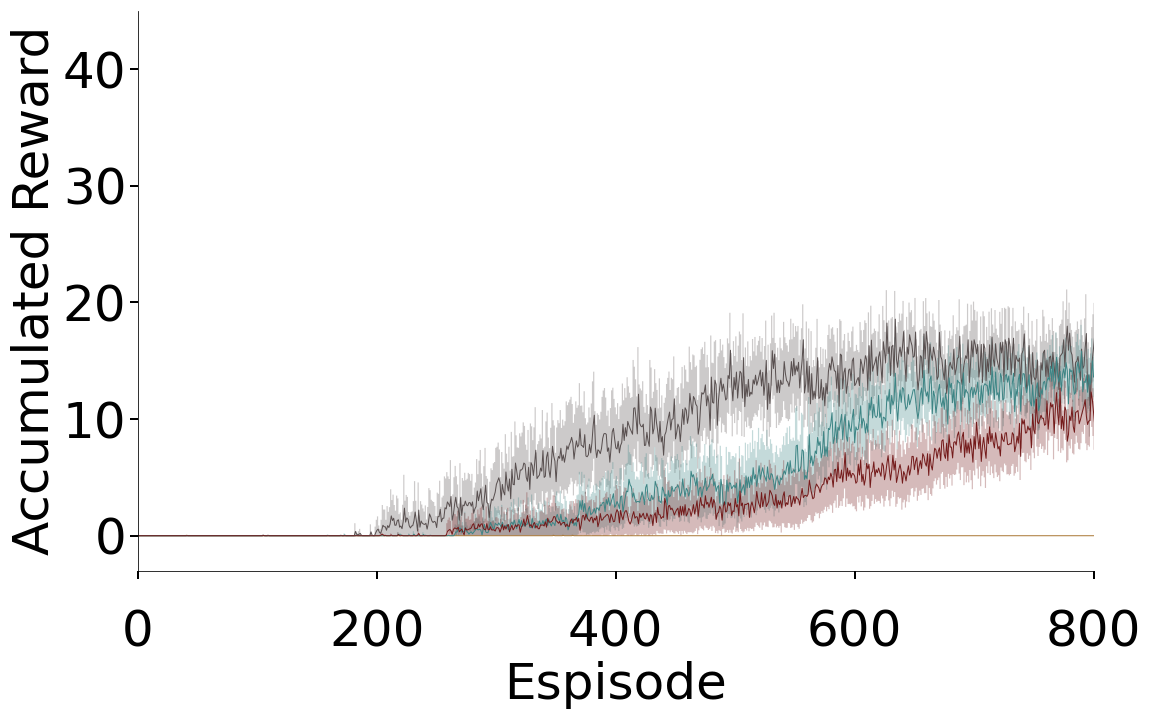}
(c) { DRIFT} environmental condition\\
\caption{\bf The effects of the of the co-agent temporal representation on early task performance. We see significant improvements from learning occur within the first 600 episodes. We compare learning co-agents coupled to an Expected Sarsa learning agent over the first 800 episodes (8000 steps) in the (a) fixed, (b) random, and (c) drift conditions. Lines show the means over 30 independent runs; shaded regions show the 95\% confidence interval. Legend indicates representations used, with TCT $a$ denoting decay $e^{-at}$.}
\label{fig:rep-fixed-gamma-comparison}
\Description{Each graph shows accumulated agent reward plotted against episode number for accumulation and countdown GVFs, under the three environmental conditions. Agents using the Bit Cascade representation are the first or tied for first to learn with each GVF and in each environmental condition, and reach higher early performance in the fixed environmental condition. Agents using the Bias representation earn reward only with the Accumulation GVF under the fixed environmental condition, and zero reward in all other cases. Agents using the TCT representations learn to gain reward in all cases, tying with the Bit Cascade agents with the Accumulation GVF under the random and drift conditions, and otherwise more slowly than the Bit Cascade agents.}
\end{figure}

\begin{figure}[!t]
\centering
\includegraphics[width=0.45\textwidth]{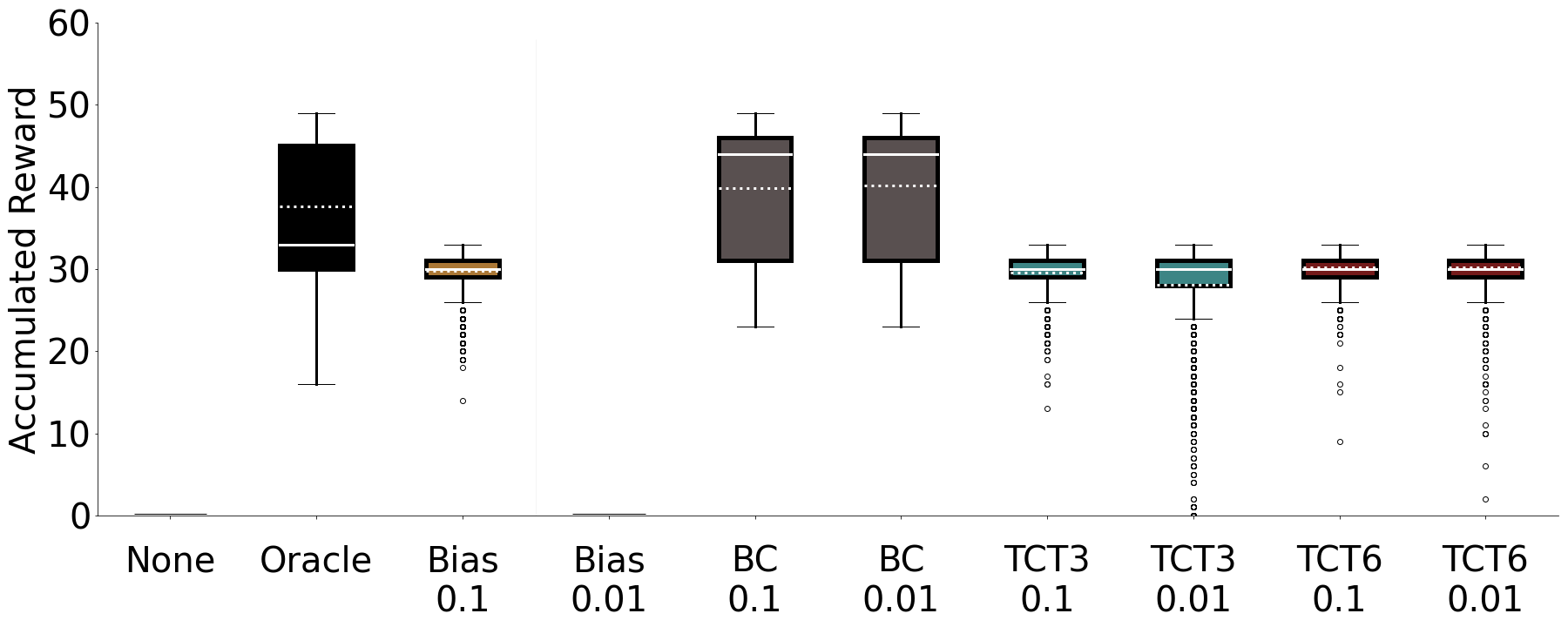}\\ 
(a) { FIXED} environmental condition\\
\ \\
\includegraphics[width=0.45\textwidth]{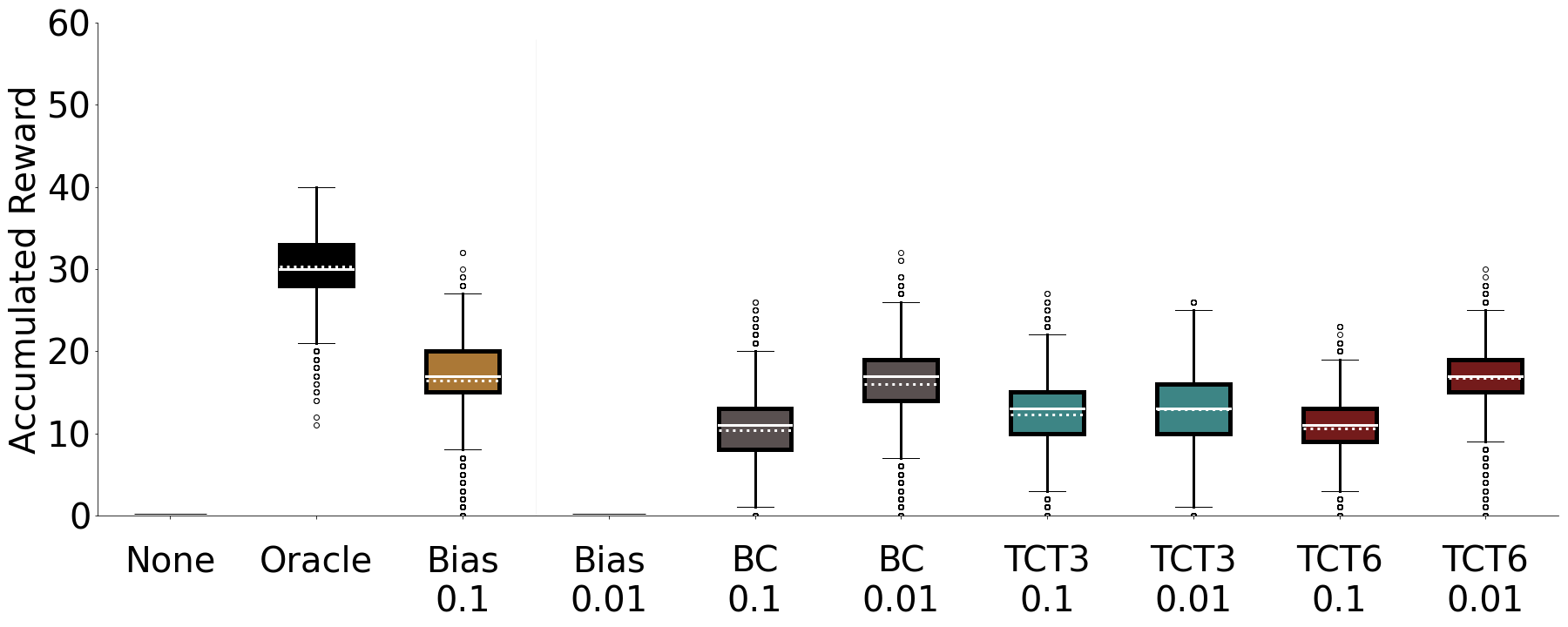}\\
(b) { RANDOM} environmental condition\\
\ \\
\includegraphics[width=0.45\textwidth]{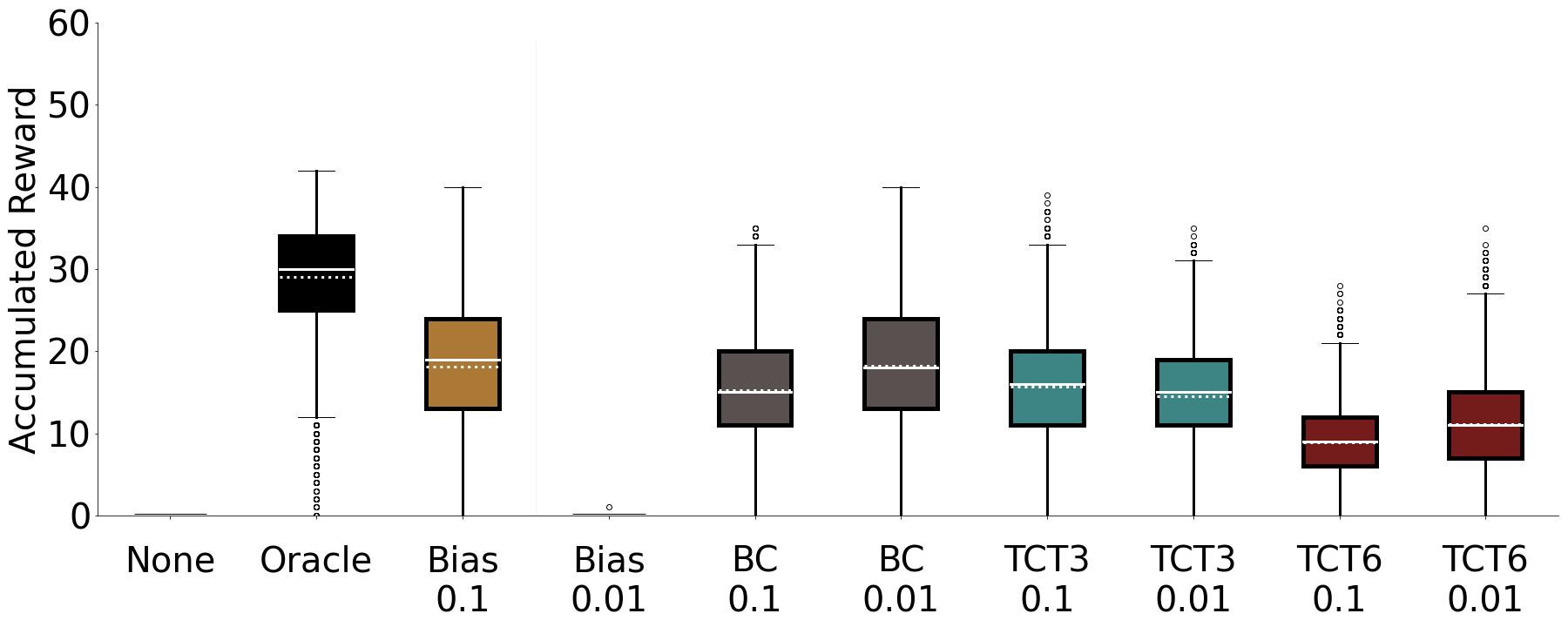}\\
(c) { DRIFT} environmental condition
\caption{Plots of accumulated reward for the last 1000 episodes with a countdown GVF across environment conditions (a--c) and temporal representations. Across all environment conditions and most temporal representations, the co-agent performance approaches that of the oracle, and is substantially better than having no co-agent. Plots show the median (solid white line), mean (dotted white line), 1st and 3rd quartile (box sections), and outliers (dots beyond the whiskers, which extend $\pm 1.5\times$ the interquartile range), over 30 independent trials. For learning co-agents, the name indicates the time representation and its parameter (if any), followed by the GVF learning rate.}
\Description{Box plots showing mean, median, first and third interquartile ranges and outliers. Results for fixed, random, and drift conditions are displayed in a vertical stack. The x axis of each is the name of the GVF co-agent and associated temporal representation. They are None, Oracle, Bias 0.1, Bias 0.01, BC 0.1, BC 0.01, TCT3 0.1, TCT3 0.01, TCT6 0.1, TCT6 0.01. The y axis is labelled accumulated reward, and ranges from 0 to 60.

The "None" co-agent is unable to achieve reward in any environment.

The oracle, and BC based agents perform best in the fixed environment, with approximately equal medians and interquartile ranges. The oracle co-agent has a slightly lower mean. The Bias 0.1, TCT3 0.1, TCT6 0.1, TCT3 0.01 and TCT6 0.01 have similar medians and interquartile ranges, but there are more outliers on the low performing tail of TCT3 0.01 and TCT6 0.01. Bias 0.01 fails to obtain reward.

The oracle performs best in the drift condition. The Bias 0.01 and BC 0.01 perform similarly, and are the next best performing agents. Bit Cascade 0.1, TCT3 0.1, and TCT3 0.01 also demonstrate similar performance. TCT6 0.01 outperforms TCT 0.1 by a small margin, and Bias 0.01 fails to obtain reward.}
\label{fig:rep-comparison-boxplot}
\end{figure}

Figure \ref{fig:nexting-short-isi-comp} shows that the choice of temporal representation affects predictions made by the co-agent, including their timing and degree of aliasing at different points in an ISI's span with respect to token generation thresholds. We here focus on performance differences induced by these choices. We ran sweeps of 5000 learning episodes each of 1000 steps in length, across agent-co-agent pairs, in the fixed (ISI 8 with 2 hazard steps), random (ISI is in the range by 5 to 10 steps, selected independently after each hazard), and drift conditions (ISI changes by -1 to 1 steps after each hazard). Early learning differences in agent performance for Pavlovian signalling co-agent / control learning agent combinations (average accumulated episodic reward) are shown in Fig. \ref{fig:rep-fixed-gamma-comparison}, while asymptotic learning performance across representations and with the oracle co-agent is shown in Fig. \ref{fig:rep-comparison-boxplot}. Of note, these results can be well considered with respect to the different forms of aliasing in our representations for accumulation and countdown predictions, as seen in Fig. \ref{fig:nexting-short-isi-comp}. 

First, we found that, unsurprisingly, Pavlovian signalling on the part of the co-agent was not just a benefit for solving the Frost Hollow tasks, but essential---agents without a co-agent were unable to obtain reward (Fig. \ref{fig:rep-comparison-boxplot}, leftmost column). In the Fixed condition, it is possible to obtain a maximum accumulated reward of 50 per episode by earning one point every second hazard cycle. The Oracle co-agent is unrealistically strong in that it directly observes the time to the upcoming hazard instead of learning it, and learning agents connected to it are often the top performers. The remaining columns in the figure present a learning co-agent with a learning agent, and we find that, with the exception of one Bias unit co-agent, all are able to consistently obtain reward in all three environmental conditions. The Bias-0.1 co-agent is interesting in that it is competitive with the other learning co-agents in all environmental conditions. Although the feature vector output by the time representation is the same in all states, the GVF does contain a weight parameter that is updated on each timestep, and with an appropriately tuned learning rate it can oscillate across the fixed threshold to output useful Pavlovian signals for the learning agent to act on (c.f., Fig. \ref{fig:nexting-short-isi-comp}d). 
Paired with a countdown GVF question, the Bias representation with a learning rate of 0.01 was unable to track the stimulus, and obtained virtually no reward in the environment, except for a single reward in one of 30 drift trials. Paired with an accumulating GVF, it obtained approximately 29.937 reward per episode in the fixed condition, but only about 0.0019 and 0.0092 reward per episode in the random and drift conditions. Other representations performed comparably, indicating that the representation was less of a factor than learning rate and environmental condition on overall agent performance.

Further, as shown in Figs.~ \ref{fig:rep-fixed-gamma-comparison} and \ref{fig:rep-comparison-boxplot} (in terms of accumulated episodic reward for 1000 step episodes), we found that early learning and asymptotic performance varied across GVF types and representations, and in best cases approached or equalled the performance of agents partnered with an oracle co-agent (Fig. \ref{fig:rep-comparison-boxplot}). When partnered with a co-agent, we found evidence of reward acquisition by the agent as early as 200--600 episodes, requiring roughly 600 episodes to approach asymptotic performance; early learning as shown in Fig.  \ref{fig:rep-fixed-gamma-comparison} was indicative of longer-term performance; GVF learning was even faster---co-agent predictions approximating the ideal return were easily learned in less than an episode (<500 steps, or 50 ISI examples), and found to adapt to track the ISI in the drift condition while maintaining expectations of the envelope of ISI lengths in the random condition. GVF-based features were found to be stable and robust, which we suggest allowed consumption by the control learning agent without problematic time and data requirements requisite to fully learned communication setting (c.f., \cite{lazaridou2020}). 
We also studied the effect of varying heat capacity and of different linear control learning algorithms; the choice of control algorithm (e.g., changes to Q-learning or Sarsa) appeared to have only a small effect on performance across conditions compared to  differences induced by other parameter or representation choices.

\section{Virtual-Reality Results}

\begin{figure*}[!th]
\centering
(a) \includegraphics[width=0.9\linewidth]{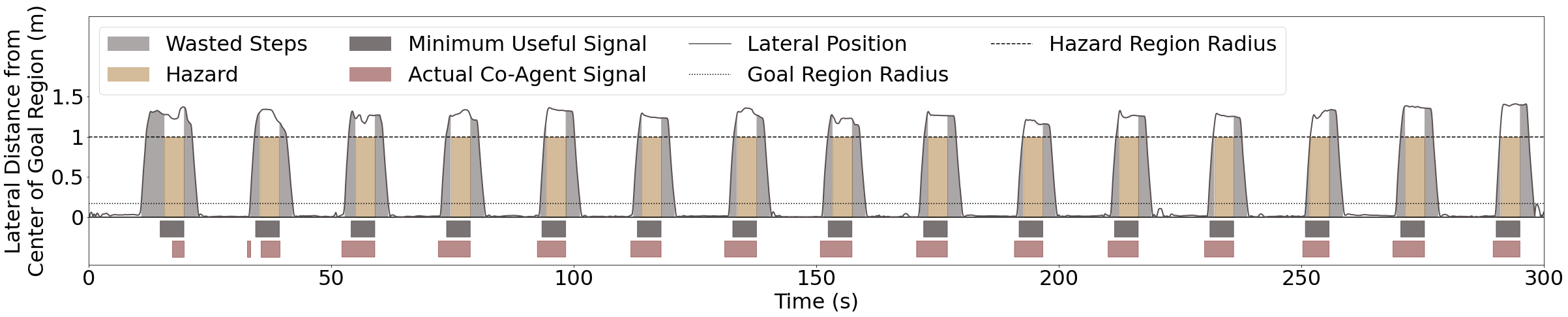}\\
(b) \includegraphics[width=0.9\linewidth]{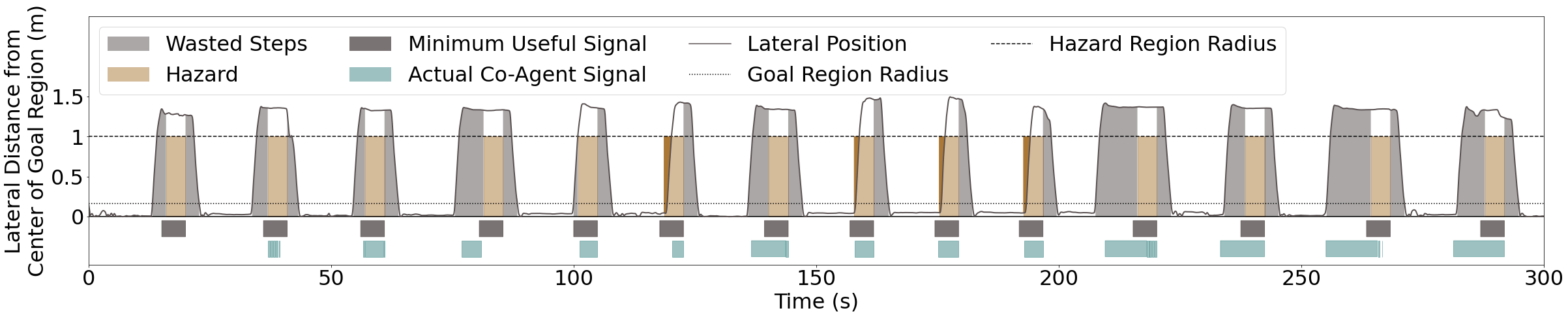}
\caption{Co-agent rapidly learns and gains utility during human interaction. Examples of a typical simple trial (a, fixed condition with TCT co-agent, red) and a typical challenging trial (b, random condition with BC co-agent, blue). {\em Wasted steps}: any time the participant was outside the heat region while the hazard pulse was inactive. {\em Minimum Useful Signal}: the signal giving the participant exactly 0.89s lead-time before the hazard begins, calculated according to participant exit velocity.}
\Description{Two panels are show stacked vertically representing trials in the virtual reality Frost Hollow environment. The both panels are graphs with an x-axis label of "Time in seconds" and a y axis label of "Lateral distance from center of goal region in meters". The range of the x axes are from 0 to 300 seconds, and the range of the y axes are from 0 to 1.5 meters. The top graphic contains a set of 15 hazards (vertical bars of magnitude 1 and width of 4 seconds) distributed on the x axis at a regular interval. The lower graphic displays 14 hazards at irregular spacing and a width of 4 seconds. The minimum useful signal from the co-agent is plotted immediately below the hazards. The co-agent signal is plotted below the minimum useful signal. }
\label{figure:h-a-example-trajectory-simple}
\end{figure*}

As a further comparison, we studied the interactions between Pavlovian signalling co-agents (BC and TCT0.3 representations) with a more advanced control learning agent: a human participant. {\em Importantly, the two co-agents were adapted directly from the abstract domain with only minimal changes to account for differences in ISI span.} We present a brief summary of our findings here.

The participant engaged with the VR Frost Hollow domain over the course of ten sessions. Each session had nine randomly ordered 5-minute trials, with the participant either working with a BC or TCT accumulation GVF co-agent, or without a co-agent, in the fixed, random and drift conditions (roughly 10 hours of participant experimental time). The initial time between hazards (ISI) was set to 20 seconds of real time (with 8 ms agent timesteps).
To make identification of individual trials in the fixed and drift conditions more challenging and to better cover the space of ISI lengths, starting ISI was set differently for each trial.
For this case study we worked with a single participant (male, age 40). Due to COVID-19 limitations in place for the duration of this work, we were unable to recruit external participants for this study, as intended and as per our approved human research ethics protocol for this work. {\em Our participant was thus a member of the study team}; we mitigated the disadvantages of this recruitment choice through trial order randomization, hiding the design of the protocol from the participant, and removing the participant from participation in the quantitative and qualitative analysis of results.

To gain an intuitive sense of how the participant engaged with the co-agent, Fig. \ref{figure:h-a-example-trajectory-simple} presents two specific trials in detail: a typical ``simple'' trial (fixed ISI condition, paired with the TCT co-agent), and a typical ``challenging'' trial (random ISI condition, BC co-agent). Beginning with the simple trial, we saw the co-agent providing a useful (though inconsistent) signal beginning on the second hazard pulse, and reliably thereafter. The amount of wasted steps leading up to the hazard pulse diminished to a narrow margin as the trial progressed. In the challenging trial we saw the co-agent give its first useful signal only by the time of the fourth pulse. Until this time, the participant had been using their internal timing to determine when to leave the heat region in advance of the signal. The co-agent was then unable to give reliable signals for the next few pulses (pulses 5, 6, and 8--10), and the participant was hit by the hazard several times. By the 11th pulse, the participant resumed reliance on their internal timing to leave the heat region in advance of the co-agent signal, wasting many steps in order to avoid the hazard. Overall, and conversely to the abstract domain, we found the participant relied more on the TCT co-agent, with aliasing of the temporal representation providing a degree of advance notice that was reported and measured to be useful in human decision making (please see Brenneis et al. \cite{brenneis2021} for more detail).

\section{Discussion}
\label{sec:discussion-multitoken}

At a high level, the results in this work support the use of Pavlovian signalling as a lens to study certain agent-agent relationships. As identified by Pilarski et al. \cite{pilarski2017}, it is possible to frame dyadic partnerships between agents in terms of the agency and capacity of the parties engaged in the interaction; while capacity of a partnership might be limited by having reduced agency by one of the parties (here the co-agent), the simplicity of that partner provides the opportunity for fast learning of its behaviour by the other party \citep{pilarski2017,pilarski2012}. This is the case with Pavlovian signalling: we see that as a main benefit {\em useful co-agent predictions could be learned rapidly from a blank slate (less than 500 steps in the abstract domain, or less than a minute in the VR domain)} (Figs. \ref{fig:rep-fixed-gamma-comparison} and \ref{figure:h-a-example-trajectory-simple}), and could continually adapt during deployment; when made into a low-bandwidth token, predictions could be used for policy learning by a control learning agent in both abstract and VR domains without substantive changes to the co-agent, {\em and approached oracle-level performance} (Fig. \ref{fig:rep-comparison-boxplot}).

In addition to rapid co-agent learning, and the learnability of co-agent signals by the main agent, we see evidence that there are further benefits to Pavlovian signalling in that it assumes nothing about the internal structure or mutability of the control agent; human and machine agents alike can approach co-agent signals from an ungrounded perspective. Further, there is no reason to limit the co-agent to only observations of the environment as we have done in these studies; it is natural to think that information pertaining to the agent and its adaptability can also be used as inputs to the co-agent, further increasing its ability to make relevant predictions and to begin to build what has previously been termed as communicative capital \citep{pilarski2017}. As such, Pavlovian signalling appears to be a viable stepping stone between between fixed agent-agent signalling and bidirectionally learned signalling relationships.

As described by Scott-Phillips \cite{scottphillips2014,scottphillips2009}, there is a route to progress from grounded signals to what is termed ostensive-inferential communication. We believe Pavlovian signalling can provide a functional bridge towards this more natural, expressive form of communication. Studies by others have provided one view into this pathway, through a bidirectional learning relationship wherein the agent (a human) made certain things visible for co-agent learning, and the co-agent subsequently learned when and how to make its Pavlovian signalling tokens visible to the agent \cite{pilarski2019}. We suggest that tokens created in Pavlovian signalling with GVFs have the beneficial property that they are constructivist in nature, and so might be created by a co-agent autonomously as opposed to specified by an external designer. Tokens are for the sender grounded in co-agent-centric (subjectively specified) GVF question parameters cumulant $C_t$, time scale $\gamma_t$, and policy $\pi$ and also in the mapping approach, in this case is parameterized by the threshold value $\tau$ used in token generation. For our specific case, this 4-element tuple $\lbrace C_t, \gamma_t, \pi, \tau \rbrace$ is the grounding of the token, and it can be fully subjective to the sending agent and not require connections to an objective frame of reference. 

\vspace{1em}

\begin{tcolorbox}[colback=yellow!7!red!5!white,colframe=yellow!35!red!25!white]
\textbf{Pavlovian signalling as formalized in this work} is a process wherein learned, temporally extended predictions in the form of general value functions are mapped via a fixed threshold to Boolean tokens intended for receipt by a decision-making agent, where  tokens are grounded for the sender in the cumulant $C_t$, time scale $\gamma_t$, policy $\pi$ and threshold $\tau$ of their computational precursors.
 \end{tcolorbox}

\newpage

\section{Conclusions}
In this work, we contributed a concrete definition and exploration of Pavlovian signalling as implemented via processes of GVF learning. Our findings, while preliminary in that they are derived from a single human-agent case series and related fundamental agent-agent experiments, suggest that {\em Pavlovian signalling by a co-agent can be learned very rapidly in single-shot real time deployment}, improves temporal decision-making of receiving human and machine control-learning agents, and opens a number of future avenues for using GVF learning in this way to augment, adapt, and potentially enhance human and machine perception, action, and cognition. When presented with information from a machine co-agent, we observed changes in both the timing of decisions of both human and machine agents (behavioural change, differences in sensorimotor trajectories, and reaction times) and also the quantitative outcomes of the timing task (score, number of mistakes).

Future study is needed with larger environments or continual learning settings with distractors, and in tasks that more fully blend both time and space. In summary, we believe there is great opportunity for using Pavlovian signalling to understand agent-agent signalling and communication in complex tasks that unfold in both time and space, and as a way-point between hand-designed communication interfaces and full machine communication learning.

\begin{acks}
The authors thank Matt Botvinick, Kory Mathewson, Michael Bowling,  Ola Kalinowska, Nathan Wispinski, Andrew Bolt, Alexander Zacherl, Richard Sutton, Kevin McKee, and Drew Purves, for many helpful conversations, discussions, and technical insights regarding this work. University of Alberta collaboration on this work by PMP, ASRP, and AW was supported by the Canada CIFAR AI Chairs program, the Canada Research Chairs program, the Natural Sciences and Engineering Research Council (NSERC), and the Alberta Machine Intelligence Institute (Amii). We also thank the authors and maintainers of Launchpad, which we used for our experiments~\cite{launchpad2021}. 
\end{acks}


\bibliographystyle{ACM-Reference-Format} 
\bibliography{main}


\end{document}